\newcommand{\CLASSINPUTtoptextmargin}{19.1mm}
\newcommand{\CLASSINPUTbottomtextmargin}{19.1mm}
\newcommand{\CLASSINPUTinnersidemargin}{19.1mm}
\newcommand{\CLASSINPUToutersidemargin}{19.1mm}
\documentclass[conference]{IEEEtran}
\IEEEoverridecommandlockouts
\usepackage{cite}
\usepackage{amsmath,amssymb,amsfonts}
\usepackage{graphicx}
\usepackage{textcomp}
\usepackage{xcolor}
\usepackage{svg}
\usepackage{url}
\usepackage{booktabs}
\usepackage{multirow}
\usepackage{graphicx}
\usepackage{rotating}
\usepackage{array}      
\usepackage{makecell}   

\renewcommand{\arraystretch}{0.85}

\def\BibTeX{{\rm B\kern-.05em{\sc i\kern-.025em b}\kern-.08em
    T\kern-.1667em\lower.7ex\hbox{E}\kern-.125emX}}

\usepackage{algorithm}
\usepackage{algpseudocode}

\usepackage{fancyhdr}
\begin{document}
\pagenumbering{arabic}

\title{\vspace{8mm}TreeIRL: Safe Urban Driving with Tree Search and Inverse Reinforcement Learning
\thanks{$\dagger$ -- work done while at Motional. }}

\author{\IEEEauthorblockN{
  Momchil S. Tomov, 
  Sang Uk Lee,  
  Hansford Hendrago,  
  Jinwook Huh, 
  Teawon Han, 
  Forbes Howington, \\ 
  Rafael da Silva,  
  Gianmarco Bernasconi, 
  Marc Heim,  
  Samuel Findler,  
  Xiaonan Ji, 
  Alexander Boule,  \\  
  Michael Napoli, 
  Kuo Chen, 
  Jesse Miller$^\dagger$, 
  Boaz Floor, 
  Yunqing Hu 
  }
   \IEEEauthorblockA{\textit{Motional AD Inc.}\\
  \texttt{\{momchil.tomov,sang.lee,hansford.hendargo,boaz.floor,alex.hu\}@motional.com}}
}


\maketitle

\thispagestyle{fancy}


\begin{abstract}
We present TreeIRL, a novel planner for autonomous driving that combines Monte Carlo tree search (MCTS) and inverse reinforcement learning (IRL) to achieve state-of-the-art performance in simulation and in real-world driving. The core idea is to use MCTS to find a promising set of safe candidate trajectories and a deep IRL scoring function to select the most human-like among them. We evaluate TreeIRL against both classical and state-of-the-art planners in large-scale simulations and on 500+ miles of real-world autonomous driving in the Las Vegas metropolitan area. Test scenarios include dense urban traffic, adaptive cruise control, cut-ins, and traffic lights. TreeIRL achieves the best overall performance, striking a balance between safety, progress, comfort, and human-likeness. To our knowledge, our work is the first demonstration of MCTS-based planning on public roads and underscores the importance of evaluating planners across a diverse set of metrics and in real-world environments. TreeIRL is highly extensible and could be further improved with reinforcement learning and imitation learning, providing a framework for exploring different combinations of classical and learning-based approaches to solve the planning bottleneck in autonomous driving.
\end{abstract}


\section{Introduction}
\label{sec:introduction}

\begin{figure*}
    \centering
    \includegraphics[width=1\textwidth,trim={0 50 0 0},clip]{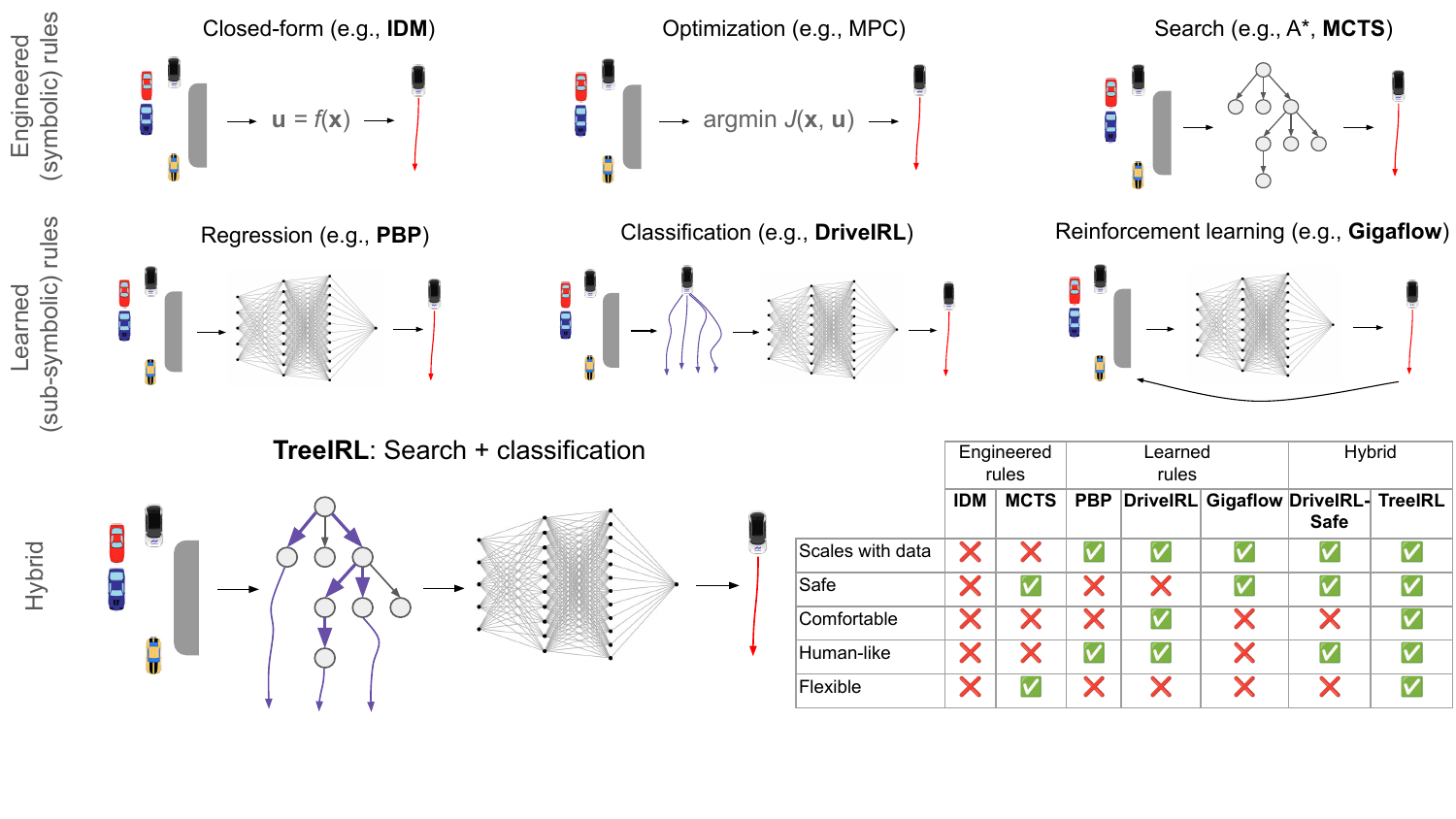}
    \caption{Landscape of motion planning approaches and summary of our results.}
    \label{fig:intro}
\vspace{-3mm}
\end{figure*}

Human-level planning and decision making remain the holy grail of autonomous driving, promising to make transportation safer, cheaper, and more accessible to everyone. Mirroring broader trends in artificial intelligence, classical approaches to motion planning that explicitly codify the rules of driving in symbolic form~\cite{lavalle2006planning,paden2016survey} have given way to approaches based on machine learning (ML) that learn the rules of driving from data and represent them implicitly in sub-symbolic form~\cite{kuutti2020survey,karnchanachari2024towards}. While similar developments have fueled remarkable success in other domains -- such as image processing~\cite{krizhevsky2012imagenet}, language processing~\cite{vaswani2017attention}, game play~\cite{mnih2015human}, and even perception and prediction for self-driving cars~\cite{caesar2020nuscenes,Zhou_2018_CVPR, Lang_2019_CVPR, vora_2020_cvpr, phan-minh_2020_cvpr, yin_2021_cvpr, chen_2021_neurips, liang_2020_eccv} -- ML-based planners have struggled to live up to their promise, raising questions about their utility~\cite{dauner2023parting}.

Some classical approaches frame the planning problem as tree search over an appropriate state space~\cite{dolgov2010path,paden2016survey,sunberg2017value,heim2025lab2car}. These approaches can ensure safety with explicit checks and 
can flexibly find solutions in a wide variety of on-road scenarios~\cite{dolgov2010path}. However, they can occasionally produce uncomfortable and unnatural behavior, something difficult to remedy with data due to their non-differentiability and small number of parameters.

ML-based approaches often frame the planning problem as trajectory regression or classification and are trained on many hours of human driving~\cite{Bojarski_2016_arxiv,Codevilla_2019_ICCV,Bansal_2019_RSS,scheel2022urban,phan2023driveirl}. These approaches scale with data, producing increasingly human-like behavior as their many parameters are refined with training.
However, they can struggle to ensure safety, as safety-critical cases are rarely encountered on the road and in the training data~\cite{zhou2022long}. Similarly, they can struggle to generalize flexibly to scenarios outside of the data distribution~\cite{kuutti2020survey}.

In this work, we propose TreeIRL, a hybrid approach that combines classical tree search with a learned classifier to yield a planner that is safe, comfortable, and human-like (Fig.~\ref{fig:intro}). The main idea is to repurpose \textit{Monte Carlo tree search (MCTS) as a trajectory generator}: instead of computing a single immediate next action (e.g., control command), as is usually done, MCTS generates a set of possible future action sequences (i.e., trajectories). These candidate trajectories are fed into a scoring function trained on expert human driving using inverse reinforcement learning (IRL). 

MCTS efficiently explores the trajectory space at decision time to home in on trajectories that are generally appropriate for the current situation, effectively selecting the behavior mode (e.g., slow down, accelerate, stop). The IRL scorer then takes advantage of the variability around that mode to further refine behavior by selecting the most human-like trajectory for execution (e.g., slow down gradually vs. abruptly). Intuitively, this division of labor delegates safety and progress to MCTS and comfort to IRL, although we find empirically that the two complement each other across multiple metrics to yield a system that is greater than the sum of its parts.

We evaluate TreeIRL against classical and state-of-the-art motion planners on challenging urban driving scenarios in simulation and in the real world. We find that TreeIRL strikes a balance between safety, comfort, and progress that achieves the best overall performance. Our contributions are as follows:

\begin{enumerate}
    \item We show how MCTS can be repurposed as a trajectory generator and combined with IRL to reap the benefits of both classical and ML-based motion planning.
    \item We show how to optimize the latency of the resulting system in the context of a full autonomous vehicle (AV) stack, making it suitable for real-world deployment without sacrificing driving performance.
    \item We provide the first real-world demonstration of motion planning based on MCTS in dense urban traffic.
    \item We demonstrate the superiority of TreeIRL by comparing it against vanilla MCTS and other state-of-the-art planners in large-scale evaluations both in simulation and in real-world urban driving.
    \item We highlight the need for considering a diverse set of metrics and for performing on-road evaluations to address the simulation-to-reality (sim-to-real) gap.  
\end{enumerate}

\section{Related work}
\label{sec:related_work}

\textbf{Classical approaches} to the motion planning problem rely on explicit hand-crafted rules that dictate driving behavior at decision time~\cite{lavalle2006planning,paden2016survey}. 
The seminal intelligent driver model (IDM)~\cite{treiber2000congested} directly computes the command acceleration to maintain a safe distance to a lead vehicle, ensuring collision-free behavior for simple 1-D adaptive cruise control (ACC). 
Balancing more complex objectives (e.g., reaching a goal while maintaining kinematic feasibility and avoiding obstacles) in higher-dimensional environments can be framed as a continuous optimization problem and solved, for example, using model-predictive control (MPC)~\cite{qin2003survey,rajamani2006vehicle,liu2017path}. Alternatively, the search space can be discretized and explored using graph search methods such as A*~\cite{hart1968formal,dolgov2010path} or MCTS~\cite{browne2012survey,sunberg2017value}. Search and optimization can be combined -- for example, A* can find a collision-free path to the goal, while MPC can compute a dynamically feasible trajectory to track it~\cite{heim2025lab2car}.

Classical approaches can produce reasonable driving behavior while providing interpretability and, under certain conditions, safety guarantees. Search methods in particular have the potential to handle a wide variety of scenarios as they can flexibly reason and find solutions even in unusual situations (e.g, navigating a crowded parking lot after a football game). However,  handcrafted rules occasionally lead to uncomfortable and unnatural behavior. Furthermore, they have to be tuned manually, which can be time-consuming and scale poorly to scenarios requiring subtly different rules, such as different geolocations or different times of day. 

\textbf{Imitation learning (IL)} -- or behavior cloning -- approaches learn the rules of driving from human demonstrations and store them implicitly in the weights of a neural network~\cite{Bojarski_2016_arxiv,Codevilla_2019_ICCV,Bansal_2019_RSS,scheel2022urban}. The promise of IL is scalability: behavior should improve with more training data. Simple regression methods such as path-based trajectory prediction (PBP) have achieved state-of-the-art performance on motion forecasting~\cite{afshar2023pbp}.

With sufficient training, IL can yield comfortable, human-like driving behavior in the majority of scenarios. However, it can struggle on safety-critical edge cases that are underrepresented in the training data (e.g., aggressive cut-ins, jaywalkers)~\cite{zhou2022long}. More generally, na\"ive IL suffers from a distribution shift problem~\cite{ross2011reduction,kuutti2020survey,de2019causal}: the discrepancy between the scenarios observed during training and test time leads to subtle deviations in behavior, which are compounded as errors accumulate in closed loop, leading to situations completely outside of the training distribution and to undefined behavior. 

\textbf{Reinforcement learning (RL)} approaches mitigate the distribution shift problem by training a driving policy to satisfy a reward function using closed-loop simulations~\cite{pan2017virtual,liang2018cirl,zhang2021end}. Through trial-and-error, the policy learns to take actions that lead to reward (e.g., making progress along the route) while avoiding punishment (e.g., colliding with other agents). By virtue of experiencing the outcomes of their own actions and a much wider variety of scenarios than typically encountered on the road -- including many collisions and the associated ``pain" -- the resulting policies are more robust to perturbations and produce safer driving than IL.

While safe and scalable, RL approaches depend heavily on the reward function and the simulation environment, which pose their own set of challenges. Designing a reward function that balances different driving objectives in a way that leads to safe, human-like driving carries many of the same difficulties as designing good classical planners~\cite{kiran2021deep}. Designing a realistic simulation environment with human-like reactive agents can prove just as difficult as solving the planning problem itself~\cite{suo2021trafficsim,cusumano2025robust}. Training with simple reactive agents can produce unnatural behaviors (e.g., no fear of rear collisions), and so can training with realistic but non-reactive agents replayed from real-world driving data (e.g., excessive fear of rear collisions)~\cite{karnchanachari2024towards}. Recently, Gigaflow~\cite{cusumano2025robust} addressed this circular dependency using self-play: the same policy that controls the AV also controls the other agents, ensuring the simulation also improves gradually over the course of training. This simple concept -- combined with massive amounts of training -- has allowed Gigaflow to surpass the state of the art on multiple autonomous driving benchmarks.

\textbf{Inverse reinforcement learning (IRL)} promises to address the reward design problem by assuming that human driving is guided by an implicit reward function, which IRL attempts to reverse engineer from human demonstrations~\cite{abbeel2004apprenticeship,huang2021driving}. In particular, DriveIRL~\cite{phan2023driveirl} uses this idea to cast motion planning as a classification problem: a trajectory generator proposes a set of candidate trajectories and a learned reward function selects the best trajectory among them. DriveIRL demonstrated comfortable, human-like driving on a real self-driving car in Las Vegas. However, it inherits some of safety issues of IL, requiring multiple takeovers by the safety driver.

\textbf{Hybrid methods} promise to combine the benefits of different approaches. In DriveIRL, ``bad'' trajectories proposed by the generator can be excluded using a rule-based safety filter~\cite{phan2023driveirl,tomov2023safety}, substantially improving safety. As another example, IL can be combined with RL to produce a policy that is human-like and robust in rare, safety-critical scenarios~\cite{lu2023imitation}. SafetyNet~\cite{Vitelli_2022_ICRA} projects an infeasible ML trajectory onto a heuristically generated set of lane-follow trajectories. Lab2Car~\cite{heim2025lab2car} uses the ML trajectory to construct a set of spatiotemporal constraints (a ``maneuver'') that capture comfort and safety, which are then satisfied by MPC.

\textbf{MCTS} offers a particular kind of hybrid that combines classical search with ML in a principled way. The landmark defeat of Go champion Lee Sedol by AlphaGo was achieved with MCTS guided by a policy trained with IL and RL~\cite{silver2016mastering}. Subsequent work combining MCTS with ML has achieved similarly groundbreaking results in a number of other domains, including other board games~\cite{silver2018general} and video games~\cite{schrittwieser2020mastering}. Recent work has demonstrated that MCTS combined with ML also holds promise in the domain of autonomous driving~\cite{hoel2019combining,chekroun2024mbappe}.

The ability of ML-guided MCTS to explore intractably large trajectory spaces at decision time can in principle lead to robust, flexible driving across a diverse set of scenarios. However, the main challenge is latency: the number of iterations necessary to find a good solution grows exponentially with the search space. While ML can significantly reduce that number~\cite{hoel2019combining}, this comes at the cost of running the ML policy at each iteration of MCTS, which can offset those gains by dramatically increasing overall latency. As a result, to the best of our knowledge, all reported applications of MCTS to self-driving cars have been evaluated only in simulation.

The key insight of our approach is that, instead of producing a single best action or trajectory, \textit{MCTS can produce a set of candidate trajectories}. These trajectories can then be scored by a reward function learned with IRL to choose the one which is most human-like. This relaxes the requirements on MCTS, allowing for a simpler version that fits into the computational budget of a real self-driving car without sacrificing overall driving performance.

\section{Theoretical background}
\label{sec:theory}

\begin{figure*}
    \centering
    \includegraphics[width=1\textwidth,trim={0 140 130 0},clip]{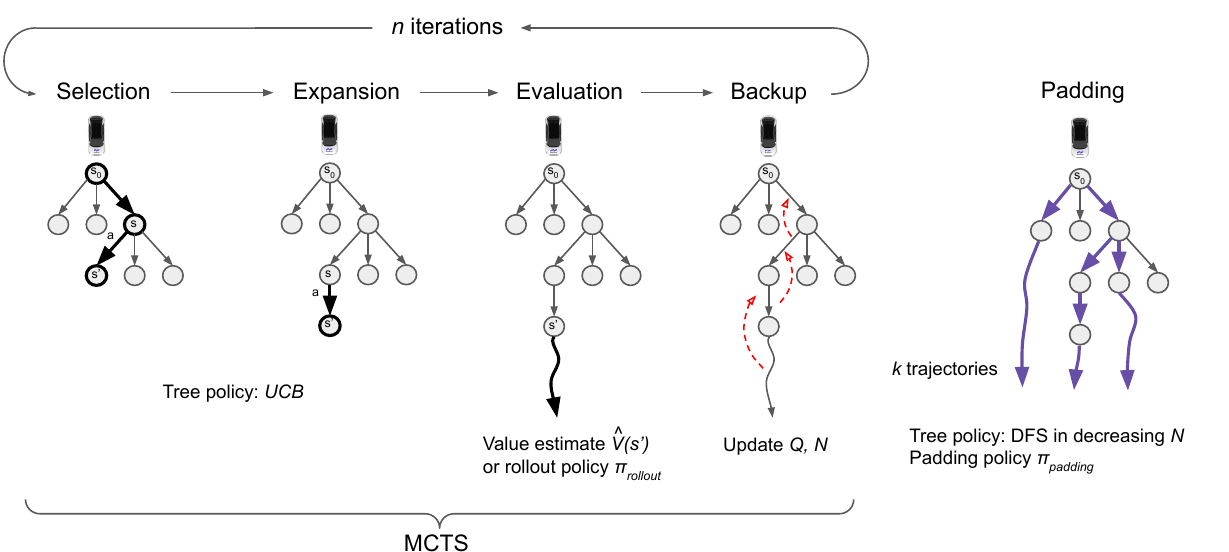}
    \caption{MCTS trajectory generator.}
    \label{fig:mcts}
\vspace{-3mm}
\end{figure*}

In this section, we review the theoretical background of MCTS.

\subsection{Markov Decision Process (MDP)}

We build on the MDP formalism developed in the RL literature~\cite{sutton2018reinforcement}. A MDP is a tuple $\mathcal{M} = \langle \mathcal{S}, \mathcal{A}, T, R, \gamma, F \rangle$ where $\mathcal{S}$ is the state space, $\mathcal{A}$ is the action space, $T : \mathcal{S} \times \mathcal{A} \to \mathcal{S}$ is the transition distribution, $R : \mathcal{S} \times \mathcal{A} \times \mathcal{S} \to \mathbb{R}$ is the reward function, and $\gamma \in (0,1]$ is the discount factor. The transition distribution $T(s' \mid s,a)$ describes the probability of transitioning to state $s'$ after taking action $a$ in state $s$. The reward function $R(s,a,s')$ describes the reward obtained by the agent after taking action $a$ in state $s$ and transitioning to state $s'$. Additionally, it is useful to introduce a termination function $F(s) \to \{0,1\}$ that denotes whether state $s$ is terminal. 

At decision time, the goal of the agent is to choose an action $a^{*}$ that maximizes its expected return (i.e., sum of future discounted rewards), starting from current state $s_0$:

\begin{equation}
\label{eq:mdp}
a^{*} = \arg\max_{a \in \mathcal{A}} \; \mathbb{E} \!\left[ \sum_{j=0}^{\infty} \gamma^{j} \, R(s_j, a_j, s_{j+1}) 
\right],
\end{equation}

where $j$ denotes the time step in the episode and the expectation is taken over the distribution of future states $s_{j+1} \sim T(\cdot \mid s_j,a_j)$ and actions $a_j \sim \pi(\cdot \mid s_j)$ resulting from the transition distribution $T$ and the agent's policy $\pi:  \mathcal{S} \to \mathcal{A}$.

\subsection{Rollout algorithms}

Rollout algorithms correspond to a class of decision-time Monte Carlo planning algorithms that solve Eq.~\ref{eq:mdp} by repeatedly sampling trajectories from the initial state $s_0$ following a \textit{rollout policy} $\pi_\text{rollout}$, simulating their outcomes, 
and choosing the action $a^*$ that produces the highest average return across the simulations. These approaches fall under the banner of model-based RL, as they require complete model of the environment -- the MDP, or an estimate of it -- in order to perform the simulations.

\subsection{Monte Carlo tree search (MCTS)}

MCTS improves on rollout algorithms by taking advantage of the fact that most trajectories will share the same initial states, which means that previously sampled trajectories can inform future trajectories (e.g., if going left consistently leads to simulated collisions, we should try something else)~\cite{browne2012survey}. 

In particular, MCTS incrementally builds a tree rooted in the present state $s_0$ in which nodes correspond to states and edges correspond to actions leading to next states (child nodes). MCTS expands the tree in multiple iterations, where each iteration consists of the following four steps (Fig.~\ref{fig:mcts})~\cite{sutton2018reinforcement}:

\begin{enumerate}
    \item Selection: starting from the root state $s_0$, follow a \textit{tree policy} that descends down the tree, balancing exploration and exploitation.
    \item Expansion: once a new state is reached, (optionally) add it to the tree as a new leaf node.
    \item Evaluation: estimate the expected return from the leaf state, either using a value function approximator $\hat{V}$ (bootstrap estimate), or by computing the simulated return using a rollout policy $\pi_\text{rollout}$ (Monte Carlo estimate).
    \item Backup: update the tree value estimates $Q(s,a)$ and visit counts $N(s,a)$ for all leaf ancestors.
\end{enumerate}

After $n$ of iterations -- often dictated by a compute budget -- the tree has $O(n)$ nodes. In the end, the best action can be chosen either as $a^* = \arg \max_a Q(s_0,a)$ or $a^* = \arg \max_a N(s_0,a)$.

\subsection{MCTS as a trajectory generator}

The main novel idea behind TreeIRL is that MCTS can be repurposed to generate a set of trajectories, i.e. promising sequences of actions, rather than a single next action $a^*$ (Fig.~\ref{fig:mcts}, right). After $n$ iterations, we perform depth-first search (DFS) from the root state $s_0$ by visiting child nodes in decreasing order of $N(s,a)$ -- that is, most popular children first. Let $\{l_1,l_2 ... l_k\}$ correspond to the first $k$ leaves visited by the DFS (the ``top $k$'' leaves). We then perform a rollout from each $l_i$ by following a padding policy $\pi_\text{padding}$ until we reach a terminal state. The resulting sequence of state-action pairs from the root state $s_0$ to a terminal state corresponds to trajectory $\tau_i$.

\section{TreeIRL}
\label{sec:tree_irl}

This section includes a technical description of the TreeIRL planner.


\begin{figure*}
    \centering
    \includegraphics[width=1\textwidth,trim={37 200 140 0},clip]{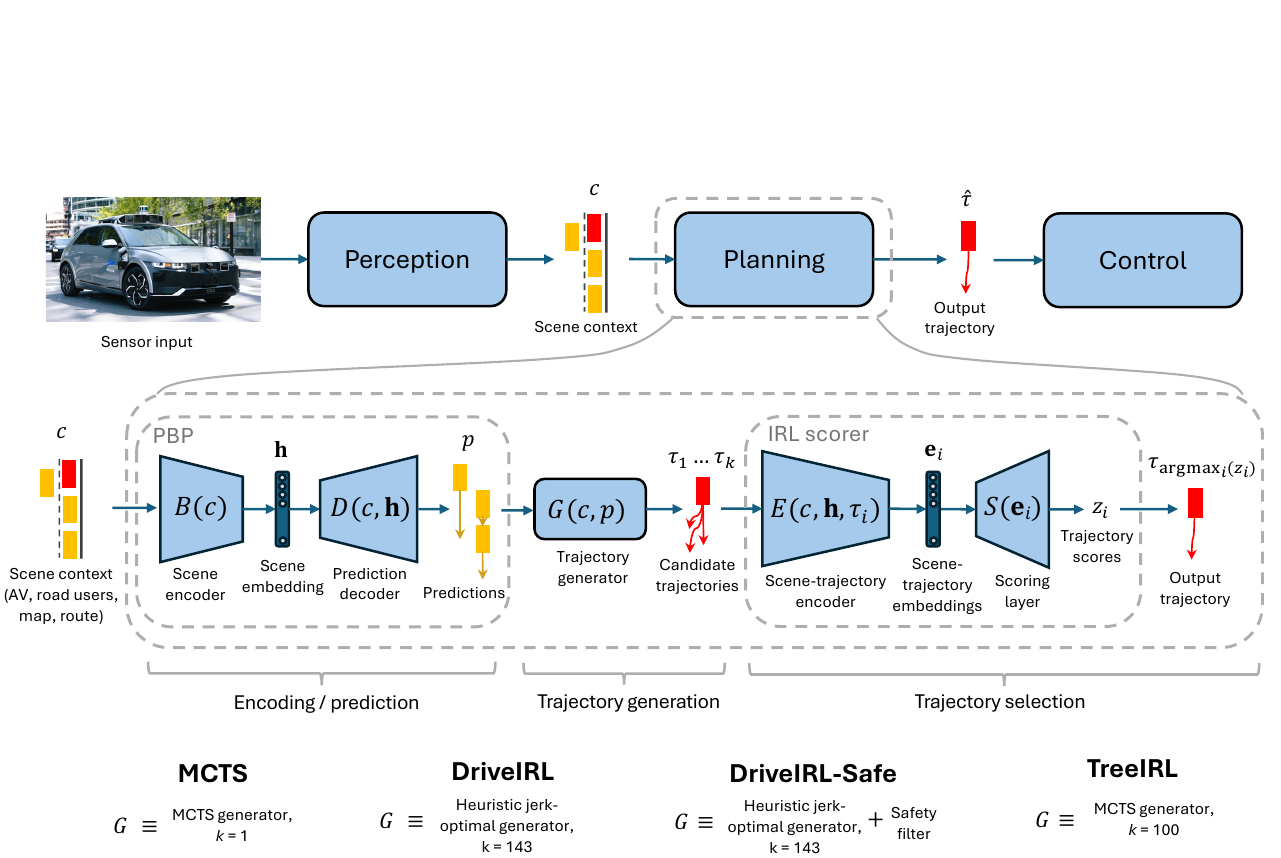}
    \caption{Planner architecture (skip connections omitted).}
    \label{fig:arch}
\vspace{-3mm}
\end{figure*}

\textbf{Planner architecture}. 
Our focus is the planning module of the AV stack (Fig.~\ref{fig:arch}, top), which is responsible both for high-level decision making (the behavior mode, e.g. ``slow down'') as well as low-level motion planning (the trajectory to follow). We assume an upstream perception module computes the scene context $c$ -- an object-oriented representation of the scene around the AV (also referred to as the \textit{ego vehicle} or simply the \textit{ego}) which includes kinematic information about the ego and the other agents, in addition to map and route information. The output of the planning module is a trajectory $\hat{\tau}$ that is passed to the downstream control module for tracking and execution.

Specifically, we use a version of the DriveIRL architecture (Fig.~\ref{fig:arch}, middle)~\cite{phan2023driveirl}, which uses a modular deep neural network that separates planning into three sub-modules:

\begin{itemize}
    \item Encoding/prediction: a scene encoder $B$ and decoder $D$ compute a scene embedding $\mathbf{h}$ and predictions $p$ for the other agents. For this module, we use PBP~\cite{afshar2023pbp}.
    \item Trajectory generation: a trajectory generator $G$ computes a set of $k$ candidate trajectories $\{\tau_1 \dots \tau_k\}$. For DriveIRL, $G$ is a heuristic generator that computes jerk-optimal trajectories that reach a handpicked set of longitudinal target offsets. It is optionally followed by safety filter that excludes trajectories likely to result in collision, a version that we refer to as DriveIRL-Safe. 
    \item Trajectory selection: an IRL scorer, consisting of a scene-trajectory transformer $E$ and a score transformer $S$, computes a score $z_i$ for each trajectory $\tau_i$, which quantifies how human-like it is.
\end{itemize}

TreeIRL uses the same planner architecture and components as DriveIRL, except for the generator $G$, which is replaced with MCTS (Fig.~\ref{fig:mcts}).


\textbf{Inference}.
During inference, the trajectory with highest score is selected on each iteration:

\begin{align*}
 \{\tau_1 \dots \tau_k\} &\sim G(c,D(c,B(c))) \\
 \hat{\tau}  &= \underset{\tau \in \{\tau_1 \dots \tau_k\}}{\arg\max} \, S(E(c,B(c),\tau))
\end{align*}

\textbf{Training}.
We pre-train PBP ($B$ and $D$) for multi-agent motion prediction, as described previously~\cite{afshar2023pbp}. We then keep those weights frozen and train the IRL scorer ($E$ and $S$; $G$ is not differentiable) on 1,360 hours of human expert driving using maximum-entropy IRL~\cite{ziebart2008maximum,wulfmeier2015maximum,phan2023driveirl}, formulated as a classification problem and optimized using a focal loss~\cite{lin2017focal}.
In particular, the probability of selecting the optimal trajectory is $P(\tau^{*}) = \frac{\text{exp}(z^{*})}{\sum_{i} \text{exp}(z_{i})}$, where $\tau^{*}$ is the optimal trajectory and $z^*$ is its score (logit). Combining a negative log-likelihood loss with a focal loss gives the following loss for a single training sample:

\[
    loss = - (1 - P(\tau^{*}))^{\gamma} \, \log P(\tau^{*}),
\]


where $\gamma$ refers to the focusing parameter, not to be confused with the MDP discount factor (also denoted $\gamma$) elsewhere. For DriveIRL, $\tau^*$ corresponds to the trajectory from $\{\tau_1 \dots \tau_k\}$ that is closest to the human expert ground truth trajectory in L2 distance (position and velocity), excluding trajectories in collision with the future trajectories of other agents. 

TreeIRL is trained in the same way as DriveIRL, except that the L2 error decays exponentially for waypoints farther into the future. We put greater emphasis on the initial waypoints since 1) they are more critical for closed-loop behavior, 2) they correspond to actions that have been more thoroughly explored and evaluated by MCTS, 3) the predictions for those waypoints are likely to be more accurate, and 4) ground truth human expert behavior (during training) is more predictable in the short term. Overall, this allows us to circumvent a number of issues that arise when training with longer trajectories, such as mode collapse on scenarios in which the expert is reacting to future events that cannot possibly be anticipated at the present (e.g., the traffic light turning green, or the lead vehicle starting to move).

We also up-weigh the velocity L2 by 5x to place greater emphasis on matching the expert speed profile, which improves following behavior and comfort in the 1-D longitudinal domain considered here. 




\textbf{Limitations of enumerative trajectory generation}.
One of the key insights behind DriveIRL is that the planning problem can be simplified when separated into trajectory generation and selection in a fashion reminiscent of generative adversarial networks~\cite{goodfellow2014generative}. In particular, the trajectory generator does not have to be particularly sophisticated, as long as it provides sufficient coverage (i.e., maximum recall). In turn, the selection mechanism -- the learned scoring function -- merely has to discriminate between ``good'' and ``bad'' trajectories (i.e., maximum precision), rather than having to generate the best trajectory from scratch.

\begin{figure}
    \centering
    \includegraphics[width=0.5\textwidth,trim={0 0 0 0},clip]{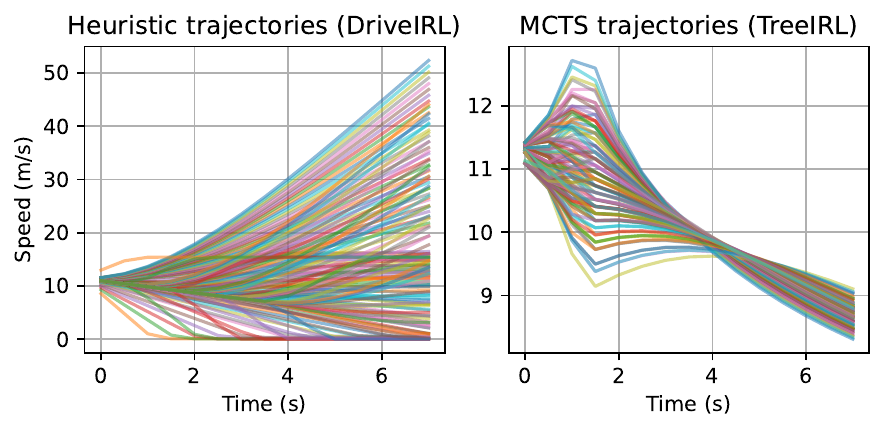}
    \caption{Example open-loop trajectories for the same scenario.}
    \label{fig:trajectory_comparison}
\vspace{-3mm}
\end{figure}

However, in practice it is challenging to design a trajectory generator that provides sufficient coverage across a diverse set of scenarios. In particular, most trajectories output by the one-size-fits-all heuristic trajectory generator of DriveIRL are inappropriate for most scenarios, either accelerating or decelerating too much (Fig.~\ref{fig:trajectory_comparison}, left). Even worse, sometimes all of the proposed trajectories are inappropriate and may even be ruled out by the safety filter, making it impossible for the scorer to find the needle in the haystack.

The key innovation behind TreeIRL is to replace $G$ with a trajectory generator based on MCTS, which focuses trajectory selection on the narrow part of the trajectory space that is behaviorally appropriate for the current scenario (Fig.~\ref{fig:trajectory_comparison}, right). For the rest of this section, we describe the technical details of the MCTS trajectory generator (Fig.~\ref{fig:mcts}).

\subsection{MDP components}

We focus on lane following and adaptive cruise control (ACC). The state and action spaces are restricted to 1-D longitudinal control along a predefined reference path, in our case corresponding to the lane centerline. As we show in the results section, even this simple setup can pose challenges for state-of-the-art approaches, particularly in real urban driving scenarios.

\textbf{State space}. Each state includes the longitudinal offset $x$, velocity $\dot{x}$, and acceleration $\ddot{x}$ of the ego vehicle and the lead agent (if there is one), as well as the time offset $t$ in the planning horizon (useful for determining termination and indexing into predictions). It also includes static information, such as the maximum longitudinal offset $x_\text{max}$ (e.g., the goal, or a red traffic light) and the speed limit $\dot{x}_\text{max}$:

\[
s = (x_\text{ego}, \dot{x}_\text{ego}, \ddot{x}_\text{ego}, x_\text{lead}, \dot{x}_\text{lead}, \ddot{x}_\text{lead}, t, x_\text{max}, \dot{x}_\text{max})
\]

We set $t = 0$ for the initial state $s_0$.

\textbf{Action space}. The action corresponds to the longitudinal jerk command, discretized into 5 possibilities:

\[
a = \dddot{x}_\text{ego} \in \{ -2, -1, 0, 1, 2 \}  \text{ m/s}^3
\]

\textbf{Transition function}. The ego, lead agent, and static portions of the next state $s' \sim T(\cdot \mid s,a)$ are computed separately. The static components are updated as:

{\small
\begin{align*}
t' &= t + \Delta t \\
x_\text{max}' &= x_\text{max} \\
\dot{x}_\text{max}' &= \dot{x}_\text{max}
\end{align*}
}

For the ego, we use a simple kinematic model that forward integrates the jerk command to produce the next ego state:

{\small
\begin{align*}
\ddot{x}_\text{ego}' &= 
    \operatorname{clip}\!\left(\ddot{x}_\text{ego} + \dddot{x}_\text{ego} \Delta t ,\; [-7 \text{ m/s}^2, 2 \text{ m/s}^2]  \right), \\
\dddot{x}_\text{ego}' &= 
    \frac{\ddot{x}_\text{ego}' - \ddot{x}_\text{ego}}{\Delta t},  \\
\dot{x}_\text{ego}' &= 
    \max\!\Big(0,\; \dot{x}_\text{ego} + \ddot{x}_\text{ego} \Delta t  + \tfrac{1}{2}\dddot{x}_\text{ego}' \Delta t^2 \Big), \\
x_\text{ego}' &= 
    \max\!\Big(x_\text{ego},\; x_\text{ego} + \dot{x}_\text{ego} \Delta t 
        + \tfrac{1}{2}\ddot{x}_\text{ego} \Delta t^2 
        + \tfrac{1}{6}\dddot{x}_\text{ego}' \Delta t^3 \Big),
\end{align*}
}

where $\dddot{x}_\text{ego}'$ is the effective jerk (not included in the state) that takes into account the acceleration clamp. We use a time step of $\Delta t = 0.5$ s. Note that driving in reverse is not allowed.

For the other agents, we use a non-reactive world model based on the (top mode) predictions $p$ from PBP:

{\small
\begin{align*}
x_\text{lead}', \dot{x}_\text{lead}', \ddot{x}_\text{lead}' = \mathtt{PREDICTED\_LEAD\_AGENT}(x_\text{ego}',p,t')
\end{align*}
}

In particular, the lead agent at time $t'$ is defined as the closest agent projected onto the reference path in front of the ego predicted to be within 2 m of the reference path at time $t'$. If there is no such agent, the lead portion of the state is left empty. While this results in a simplified non-reactive world model, it significantly reduces latency compared to a reactive world model.

All transitions are deterministic, so for convenience we can write $s' \gets T(s,a)$.

\textbf{Reward function}. The reward is a weighted sum of comfort, tracking, safety, and clearance terms, together with a buffer reward when the ego vehicle stops at a desired distance:

{\footnotesize
\begin{align}
 & R(s,a,s') = -\alpha \, cost(s') \, \text{, where} \nonumber \\
 & cost(s) \nonumber \\
 & =  w_{\text{jerk}} \,\dddot{x}_{\text{ego}}^{2} \label{eq:jerk_cost}   \\
 & + w_{\text{accel}} \,\ddot{x}_{\text{ego}}^{2}  \label{eq:accel_cost}  \\
 & + w_{\text{speed}} \, |\dot{x}_{\text{max}} - \dot{x}_{\text{ego}}|  \label{eq:speed_cost}  \\
 & - 2\,w_{\text{speed}} \,\mathbb{I}\big[|\dot{x}_{\text{max}}-\dot{x}_{\text{ego}}| < 0.5 \text{ m/s}\big] \label{eq:speed_limit_cost} \\
& + w_{\text{collision}} \,\mathbb{I}\big[ x_\text{ego} \ge x_{\text{lead}} \big] \,(\dot{x}_{\text{lead}}-\dot{x}_{\text{ego}})^{2} \label{eq:lead_collision_cost}  \\
&  + w_{\text{collision}} \,\mathbb{I}\big[x_{\text{ego}} \ge x_{\text{max}} \big] \, \dot{x}_{\text{ego}}^{2} \label{eq:station_collision_cost} \\
& + w_{\text{clearance}} \,\mathbb{I}\big[0 < x_{\text{lead}}-x_{\text{ego}} < \delta \big] \,(x_{\text{lead}}-x_{\text{ego}}-\delta)^{2} \label{eq:lead_clearance_cost} \\
&  + w_{\text{clearance}} \,\mathbb{I}\big[ 0 < x_{\text{max}} - x_\text{ego} < \delta \big] \,(x_{\text{max}}-x_{\text{ego}})^{2}
 \label{eq:station_clearance_cost} \\
& + w_{\text{stop}} \,\mathbb{I}\big[\,\dot{x}_{\text{ego}} \approx 0 \land
       (\delta \le x_{\text{lead}}-x_{\text{ego}} < 3\,\text{m}) \big] \, (\dot{x}_\text{max} - 2 \dot{x}_\text{ego}) \label{eq:lead_buffer_cost}  \\
& + w_{\text{stop}} \,\mathbb{I}\big[\,\dot{x}_{\text{ego}} \approx 0 \land
       (0 \le x_{\text{max}}-x_{\text{ego}} < 2\,\text{m} ) \big] \, (\dot{x}_\text{max} - 2 \dot{x}_\text{ego}) \label{eq:station_buffer_cost},
\end{align}
}

where $\alpha = 1/30$ is a scaling factor and $\delta$ is the clearance buffer, which we set to $\delta = 1$ m during training and $\delta = 2$ m during evaluation. Eq.~\ref{eq:jerk_cost} and~\ref{eq:accel_cost} encourage comfort. Eq.~\ref{eq:speed_cost} and~\ref{eq:speed_limit_cost} encourage (roughly) following the speed limit. Eq.~\ref{eq:lead_collision_cost} and~\ref{eq:station_collision_cost} penalize collisions. Eq.~\ref{eq:lead_clearance_cost} and~\ref{eq:station_clearance_cost} encourage maintaining a certain clearance. Finally, Eq.~\ref{eq:lead_buffer_cost} and~\ref{eq:station_buffer_cost} encourage the ego not to stop too far behind. If there is no lead agent, the terms involving the lead are set to 0.

We use the following reward weights: 
$w_{\text{jerk}}=0.05,\; 
 w_{\text{accel}}=0.2,\; 
 w_{\text{speed}}=0.1,\; 
 w_{\text{collision}}=10.0,\; 
 w_{\text{clearance}}=10.0,\; 
 w_{\text{stop}}=0.1$.
The discount factor is $\gamma = 0.99$.

\textbf{Termination function}. The episode terminates when the planning horizon $H$ is reached:

\[
F(s) = 
\begin{cases}
1, & \text{if } t \geq H, \\
0, & \text{otherwise}.
\end{cases}
\]

We use $H = 8$ s, which means that tree search and the rollouts never exceed a depth of 16.

\subsection{MCTS components}

\begin{algorithm}[t]
\caption{Monte Carlo tree search (MCTS) algorithm}
\label{alg:mcts}
\begin{algorithmic}[1]
\Function{Search}{$s$}
    \If{$F(s)$} \Comment{Termination check}
        \State \Return $0$
    \EndIf
    \State $a \gets \arg\max_a UCB(s,a)$ \Comment{Selection}
    \State $s' \gets T(s,a)$ \Comment{Transition}
    \If{$N(s,a)=0$} \Comment{Evaluation}
        \State $v \gets 
        \begin{cases}
            0  & \text{if } F(s')=1 \\
            \Call{Evaluate}{s'}  & \text{otherwise}
        \end{cases}$
    \Else
        \State $v \gets \Call{Search}{s'}$ \Comment{Recursive search}
    \EndIf
    \State $r \gets R(s,a,s') $ \Comment{Reward}
    \State $q \gets r + \gamma\, v$ \Comment{Backup}
    \State $N(s,a) \gets N(s,a) + 1$  
    \State $Q(s,a) \gets Q(s,a) + \dfrac{q - Q(s,a)}{N(s,a)}$
    \State \Return $q$
\EndFunction
\end{algorithmic}
\end{algorithm}

We use a variant of MCTS based on the AlphaGo algorithm~\cite{silver2016mastering,hoel2019combining} that can incorporate ML to guide the tree search towards more promising actions and thus drastically reduce sample complexity. In particular, a neural network $f_\theta$ parametrized by $\theta$ can take a state $s$ and action $a$ and output a policy $\pi_\theta$ and an approximate value estimate $\hat{V}_\theta$:

\[
\big( \pi_\theta(a \mid s), \hat{V}_\theta(s) \big) = f_\theta(s, a) ,
\]

where the parameters $\theta$ are learned using RL and/or IL. In our case, we can use $f_\theta$ in the MCTS trajectory generator in three distinct ways (Fig.~\ref{fig:mcts}):

\begin{itemize}
    \item as a prior $P$ in the tree policy to guide selection,
    \item as a rollout policy $\pi_\text{rollout}$ and/or value function approximator $\hat{V}$ for leaf evaluation, and
    \item as a padding policy $\pi_\text{padding}$ to generate trajectories from the top $k$ leaves in the end.
\end{itemize}

\textbf{Selection}. For the tree policy, we use the PUCTS formula~\cite{silver2016mastering}, which is an extension of the standard upper-confidence bound (UCB) algorithm~\cite{auer2002using} that uses the tree statistics $Q$ and $N$ to balance exploration -- selecting unfamiliar actions with low $N(s,a)$ -- and exploitation -- selecting rewarding actions with high $Q(s,a)$:
 
\begin{align*}    
UCB(s,a) &= \frac{Q(s,a)}{Q_{\max}} \\
&+ c_{\text{puct}} \, P(s,a) \,
\sqrt{\frac{\sum_b N(s,b) + 1}{N(s,a) + 1}} + \varepsilon \, ,
\end{align*}

where $c_\text{puct} = 1$ is the UCB scaling factor, with higher values promoting more exploration, $Q_{\max} = 1$ is a normalization factor for the action-value estimates, and $\varepsilon \sim U(0,0.001)$ is a small amount of noise added for breaking ties. When traversing the tree, actions are selected according to $a = \arg \max_a UCB(s,a)$.

In PUCTS, the prior policy $P$ guides the search by placing greater weight on actions that are \textit{a priori} promising -- that is, before simulating their outcomes. For example, at the start of the tree search, when $Q(s,a) = 0, N(s,a) = 0 \,\, \forall s\in \mathcal{S}, \forall a\in\mathcal{A}$, $P$ will dictate which action is selected first.

In our experiments, we consider two possibilities for the prior policy $P$:

\begin{itemize}
    \item the learned RL policy $\pi_\theta$, i.e. $P(s,a) = \pi_\theta(a \mid s)$, or
    \item a uniform policy $\mathcal{U}_\mathcal{A}$, i.e. $P(s,a) = 1/|\mathcal{A}|$.
\end{itemize}

\textbf{Evaluation}. There are broadly two ways to estimate the value of a newly visited state $s$. One option is to use a bootstrap estimate, i.e. the value learned using Bellman updates during training. In our case, this corresponds to the approximate value estimate from the neural network $f_\theta$:

\[
    \mathtt{EVALUATE}(s) = \hat{V}_\theta(s)\
\]

Another option is to use a Monte Carlo estimate, i.e. the return from a simulated rollout when following policy  $\pi_\text{rollout}$:

\begin{align*}
\mathtt{EVALUATE}(s) =& \sum_{j=0}^\infty \gamma^j \, R(s_j, a_j, s_{j+1})  \, \mathbb{I}[F(s_j) = 0], \\ 
& \text{where } a_j \sim \pi_\text{rollout}(\cdot \mid s_j), \; s_{j=0} = s.    
\end{align*}

We consider three alternatives for $\pi_\text{rollout}$:

\begin{itemize}
    \item The learned RL policy, i.e. $\pi_\text{rollout} \equiv \pi_\theta$,
    \item an IDM policy, i.e. $\pi_\text{rollout} \equiv \pi_\text{IDM}$, which deterministically chooses the acceleration of an IDM, and
    \item a constant speed (CS) policy, i.e. $\pi_\text{rollout} \equiv \pi_\text{CS}$, which always chooses acceleration 0 m/s$^2$.
\end{itemize}

Notice that $\pi_\text{IDM}$ and $\pi_\text{CS}$ have a different action space (acceleration, $\ddot{x}_\text{ego}$ rather than
jerk, $\dddot{x}_\text{ego}$). Since they are only used for rollouts, we simply combine them with a modified transition function that integrates acceleration instead of jerk; the state space remains unchanged.

Our full MCTS algorithm is shown in Algorithm~\ref{alg:mcts}. 
Note that tree expansion is implicit, as a new node is added for $s$ as soon as $N(s,a) > 0$.

\textbf{Padding}. We experiment with three possibilities for the padding policy $\pi_\text{padding}$ to generate trajectories from the top $k$ leaves of the resulting tree:

\begin{itemize}
    \item The learned RL policy, i.e. $\pi_\text{padding} \equiv \pi_\theta$,
    \item the IDM policy, i.e. $\pi_\text{padding} \equiv \pi_\text{IDM}$, and
    \item the constant speed policy, i.e. $\pi_\text{padding} \equiv \pi_\text{CS}$.
\end{itemize}

\subsection{State initialization}

The MCTS generator constructs the initial state $s_0$ at $t = 0$ from the following components of the scene context $c$ and predictions $p$:

\begin{itemize}
    \item kinematic \textit{ego state}: ego center, orientation, velocity, acceleration, and size (length, width) at the present time in Cartesian baselink 2D frame (origin is the rear axle center, x-direction is forward, y-direction is left),
    \item kinematic \textit{agent states}: same information for the other agents from the perception module,
    \item agent \textit{predictions}: predicted trajectories (top mode only) for the other agents from the prediction module as 8-s waypoint sequences at 2 Hz,
    \item \textit{reference path}: the reference path (in our case, the lane centerline) from the map and routing modules as a sequence of Cartesian 2D segments. 
\end{itemize}

The longitudinal components of the \textit{ego state} projected onto the reference path are used to construct the ego portion of $s_0$ ($x_\text{ego}, \dot{x}_\text{ego}, \ddot{x}_\text{ego}$). The longitudinal components of the lead agent portion of $s_0$ ($x_\text{lead}, \dot{x}_\text{lead}, \ddot{x}_\text{lead}$) are determined in the same way as for $t > 0$ (see Transition section), except using the \textit{agent states} instead of the \textit{predictions}. The maximum longitudinal offset $x_\text{max}$ is either the goal pose or the closest red/yellow traffic light at which the ego can safely stop (whichever is closest) and remains the same during the tree search. The speed limit $\dot{x}_\text{max}$ comes from the map.

Since our focus is on longitudinal control, we assume that the ego is always on the reference path -- i.e., that the lateral deviation is 0 m -- and delegate any lateral correction to downstream post-processing.

\subsection{Trajectory post-processing}

The 1-D longitudinal trajectories resulting from the MCTS generator are converted to sequences of Cartesian 2D waypoints by taking the corresponding points along the reference path. These 2-D trajectories are then passed to the IRL scorer. Finally, the top trajectory chosen by the IRL scorer is passed to the downstream post-processing system for smoothing and ensuring kinematic feasibility.

\subsection{RL network and training}

The RL network $f_\theta$ is a multilayer perceptron with two hidden layers of 256 units each for both the policy and the value function. The network is trained using Proximal Policy Optimization (PPO)~\cite{schulman2017proximal} within Stable-baselines3~\cite{stable-baselines3} using a custom vehicle-following environment. The MDP of the RL agent is the same as the MDP used in MCTS. 
PPO hyperparameters include rollout lengths of 204,800 steps, batch size of 640, learning rate of $5 \times 10^{-4}$. During training, the RL agent observes a range of traffic scenarios, including lead vehicles maintaining constant speed, high-deceleration stop events, stop-and-go patterns, sudden cut-in maneuvers, and cases without a lead vehicle.
The policy is trained for a total of 40 million steps. We find empirically that this configuration enables the RL agent to acquire stable and goal-directed behaviors across diverse driving scenarios, effectively balancing safety, smoothness, and progress toward the goal. 




\section{NuPlan experiments}

We evaluate TreeIRL in nuPlan~\cite{karnchanachari2024towards} against classical and state-of-the-art planners on scenarios based on real-world autonomous driving logs. Our focus is on lane following and adaptive cruise control (ACC). As we show, even this restricted domain poses challenges to state-of-the-art approaches.




\textbf{NuPlan simulation}.
We use the open-source nuPlan simulator~\cite{karnchanachari2024towards} to perform 10-Hz closed-loop simulations. The ego vehicle is propagated using a two-stage controller consisting of a Linear Quadratic Regulator (LQR) tracker ~\cite{liu2021simulation,varma2020trajectory} followed by a kinematic bicycle model. The other agents are replayed from the log (log-playback). This approach has been favored by other authors \cite{kothari2021drivergym,vinitsky2022nocturne,li2022metadrive,lu2023imitation} since, by definition, it produces human-like behaviors for the other agents. However, log-playback is non-reactive, which can make the simulation result unrealistic, particularly if the ego vehicle deviates too much from the log (e.g., improbable rear collisions if it is slightly slower). We mitigate this by 1) using relatively short 20-s simulations (with 4 s of history), and 2) interpreting the metrics with caution -- e.g., rear collisions are more indicative of progress and human-likeness rather than safety.

\textbf{NuPlan dataset}.
We evaluate the planners on 7000+ scenarios corresponding to $\sim$40 hours of driving data collected in the Las Vegas metropolitan area by expert human drivers manually operating the AV. The dataset covers diverse geolocations with dense urban traffic at different times of day, including the Las Vegas Strip (68.29\%), downtown (17.01\%), airport (7.10\%), and west of Strip (4.14\%) areas. Scenarios span different behaviors of the ego and the other agents, such as starting from stationary (12\%), decelerating (12\%), cut-ins (3\%), challenging cut-ins (2\%), challenging ACC (3\%), lead vehicle braking (2\%), remaining stationary (18\%), as well as nominal driving (48\%).

\textbf{NuPlan metrics}. We compute the following metrics:

\begin{itemize}
    \item \textbf{Collisions}: instances when the ego bounding box first intersects that of another agent. We only count \textit{at-fault} collisions, that is, collisions that could have been avoided if the ego had slowed down (roughly corresponding to front collisions).
    \item \textbf{Drivable area violations}: instances when the ego bounding box is more than 0.3 m outside the mapped drivable area.
    \item \textbf{Traffic light violations}: instances when the ego crosses a stop line during a red light.
    \item \textbf{Speed limit violations}: continuous measure of how often the ego exceeds the speed limit (0 = no violations, 1 = significant violations).
    \item \textbf{Time gap}: minimum time gap (i.e., projected time-to-collision with nearby agents). Small values indicate close calls.
    \item \textbf{Progress along expert route}: ego progress along the route relative to the expert ground truth. 
    \item \textbf{Comfort}: indicates whether the ego acceleration, yaw rate, and jerk remain within bounds empirically derived from the expert data: longitudinal accel $ \in [-4.05,\, 2.40]~\text{m/s}^2$, 
        absolute lateral accel $ \leq 4.89~\text{m/s}^2$, 
        absolute yaw accel $\leq 1.93~\text{rad/s}^2$, 
        absolute yaw rate $\leq 0.95~\text{rad/s}$, 
        absolute longitudinal jerk $\leq 4.13~\text{m/s}^3$, 
        absolute jerk magnitude $\leq 8.37~\text{m/s}^3$.
    \item \textbf{Min/max longitudinal jerk:} minimum and maximum longitudinal jerk (close to 0 is most comfortable).
    \item \textbf{Min/Max longitudinal accel}: minimum and maximum longitudinal accelerations (close to 0 is most comfortable).
    \item \textbf{L2 error:} average pointwise L2 distance between ego and expert trajectories.
    \item \textbf{Deceleration/acceleration delay error}: how much later the ego begins to decelerate/accelerate compared to the expert.
    \item \textbf{Max speed error}: maximum speed difference between the ego and the expert, normalized w.r.t. the maximum expert speed.
\end{itemize}

Metrics are averaged across simulations for each planner.

\subsection{Tuning MCTS}

\begin{figure}
    \centering
    \includegraphics[width=0.5\textwidth,trim={0 0 0 0},clip]{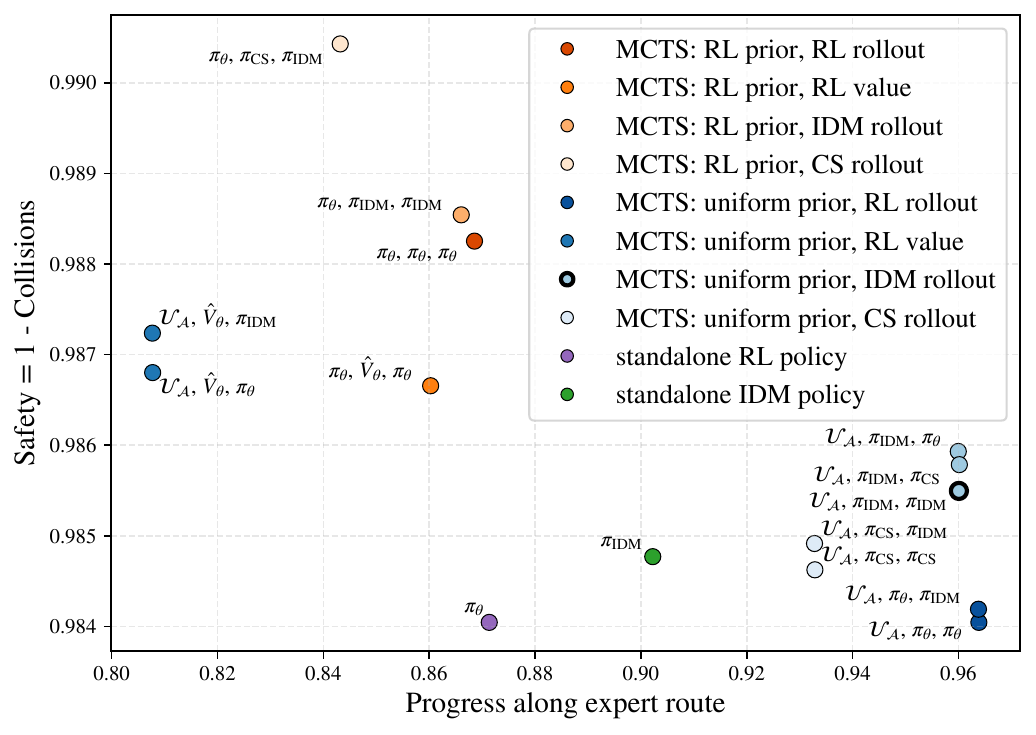}
    \caption{Comparing MCTS configurations. Highlighted configuration chosen for subsequent experiments.}
    \label{fig:mcts_variants}
\vspace{-3mm}
\end{figure}

\begin{figure}
    \centering
    \includegraphics[width=0.5\textwidth,trim={0 0 0 0},clip]{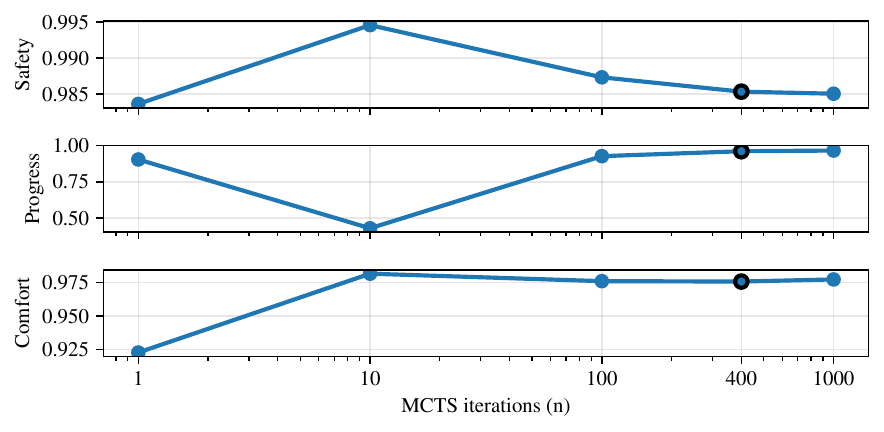}
    \caption{Number of MCTS iterations. Highlighted setting chosen for subsequent experiments.}
    \label{fig:mcts_iterations}
\vspace{-3mm}
\end{figure}

\begin{table}[t]
\centering
\caption{MCTS latency measurements}
\label{tab:mcts_latency}
\begin{tabular}{c c c c c r}
\toprule
$n$ & $k$ & $P$ & $\hat{V} / \pi_\text{rollout}$ & $\pi_{\text{padding}}$ & Latency (ms) \\
\midrule
400 & 100 & $\pi_\theta$ & $\pi_\theta$ & $\pi_\theta$ & 911.79 $\pm$ 15.26 \\
400 & 100 & $\pi_\theta$ & $\hat{V}_\theta$ & $\pi_\theta$ & 255.95 $\pm$ 4.97 \\
400 & 100 & $\pi_\theta$ & $\pi_{\text{IDM}}$ & $\pi_{\text{IDM}}$ & 45.67 $\pm$ 1.14 \\
400 & 100 & $\pi_\theta$ & $\pi_{\text{CS}}$ & $\pi_{\text{IDM}}$ & 52.32 $\pm$ 2.69 \\
400 & 100 & $\mathcal{U}_\mathcal{A}$ & $\pi_\theta$ & $\pi_\theta$ & 691.72 $\pm$ 8.73 \\
400 & 100 & $\mathcal{U}_\mathcal{A}$ & $\pi_\theta$ & $\pi_{\text{IDM}}$ & 569.49 $\pm$ 8.49 \\
400 & 100 & $\mathcal{U}_\mathcal{A}$ & $\hat{V}_\theta$ & $\pi_\theta$ & 181.38 $\pm$ 2.73 \\
400 & 100 & $\mathcal{U}_\mathcal{A}$ & $\hat{V}_\theta$ & $\pi_{\text{IDM}}$ & 48.66 $\pm$ 0.65 \\
400 & 100 & $\mathcal{U}_\mathcal{A}$ & $\pi_{\text{IDM}}$ & $\pi_\theta$ & 151.42 $\pm$ 4.03 \\
400 & 100 & $\mathcal{U}_\mathcal{A}$ & $\pi_{\text{IDM}}$ & $\pi_{\text{IDM}}$ & $^*$10.05 $\pm$ 2.79 \\
400 & 100 & $\mathcal{U}_\mathcal{A}$ & $\pi_{\text{IDM}}$ & $\pi_{\text{CS}}$ & 6.00 $\pm$ 0.14 \\
400 & 100 & $\mathcal{U}_\mathcal{A}$ & $\pi_{\text{CS}}$ & $\pi_{\text{IDM}}$ & 4.88 $\pm$ 0.10 \\
400 & 100 & $\mathcal{U}_\mathcal{A}$ & $\pi_{\text{CS}}$ & $\pi_{\text{CS}}$ & 4.93 $\pm$ 0.14 \\
0 & 1 & n/a & n/a & $\pi_\theta$ & 2.21 $\pm$ 0.05 \\
0 & 1 & n/a & n/a & $\pi_{\text{IDM}}$ & 0.03 $\pm$ 0.001 \\
\bottomrule
\addlinespace[2pt]
\multicolumn{6}{l}{\scriptsize * -- configuration chosen for subsequent experiments.} \\
\end{tabular}
\end{table}

We first compare different configurations of MCTS by taking only the top 1 trajectory (i.e., $k = 1$; Fig~\ref{fig:mcts}, bottom left), while treating the IRL scorer as pass-through. By effectively disabling the IRL scorer, we can evaluate MCTS on its own as a standalone planner, as is typically done. This allows us to explore the parameter space of the MCTS trajectory generator without having to retrain the IRL scorer for every MCTS configuration.

\textbf{MCTS configurations}. We explore different combinations of: 

\begin{itemize}
    \item prior policy $P$: RL policy ($\pi_\theta$) or uniform policy ($\mathcal{U_\mathcal{A}}$),
    \item evaluation function: RL critic ($\hat{V}_\theta$) or RL rollout ($\pi_\theta$) or IDM rollout ($\pi_\text{IDM}$) or constant speed rollout ($\pi_\text{CS}$),
    \item padding policy: RL ($\pi_\theta$) or IDM ($\pi_\text{IDM}$) or constant speed ($\pi_\text{CS}$).
\end{itemize}

We also evaluate against the standalone RL ($\pi_\theta$) and IDM ($\pi_\text{IDM}$) policies, which correspond to the special case of the MCTS generator when each of them is set the padding policy $\pi_\text{padding}$ and the number of iterations is set to  $n = 0$.

\textbf{Progress vs. safety}.
With $n=400$ iterations, using the RL policy either as a prior ($P \equiv \pi_\theta$) or as a rollout policy ($\pi_\text{rollout} \equiv \pi_\theta$) tends to produce more conservative behavior, with fewer collisions but also less progress (Fig.~\ref{fig:mcts_variants}). Results are similar when using the RL critic $\hat{V}_\theta$. In contrast, a uniform prior ($P \equiv \mathcal{U}_\mathcal{A}$) results in more aggressive behavior overall, with more progress but also more collisions. With a uniform prior, the IDM rollout policy ($\pi_\text{rollout} \equiv \pi_\text{IDM}$) is strictly better than the constant speed rollout policy ($\pi_\text{rollout} \equiv \pi_\text{CS}$).

\textbf{Iterations}.
With a uniform prior ($P \equiv \mathcal{U}_\mathcal{A}$), IDM rollout ($\pi_\text{rollout} \equiv \pi_\text{IDM}$), and IDM padding ($\pi_\text{padding} \equiv \pi_\text{IDM}$), performance plateaus around $n = 400$ iterations (Fig.~\ref{fig:mcts_iterations}).

\textbf{Latency}.
Real-world deployment makes latency a critical consideration, as the benefit of an improved planner can be lost if it is too slow to react. We compare the latency of different MCTS trajectory generators with $k = 100$ trajectories -- as they would be used with IRL -- on 100 scenarios each. Latency is measured on a Lenovo ThinkPad laptop running Ubuntu 22.04 with an Intel Core i9-10885H CPU (2.40 GHz base, up to 5.30 GHz turbo, 8 cores/16 threads) and 16 MB L3 cache, restricted to a single CPU thread (as on the car). 

Relying on the learned policy $\pi_\theta$ for the rollouts is around two orders of magnitude slower than relying on the IDM policy $\pi_\text{IDM}$ (Tab.~\ref{tab:mcts_latency}). For padding, the difference is around one order of magnitude. Our goal is to run the planner on the car at 10 Hz or more, so any latency above 100 ms is unacceptable. This effectively excludes configurations with $\pi_\text{rollout} \equiv \pi_\theta$ or $\pi_\text{padding} \equiv \pi_\theta$.

\textbf{Intermediate discussion}. The relatively conservative behavior of the learned policy $\pi_\theta$ is likely due to the low-dimensional state space, which only captures the ego and lead vehicles. Since the RL policy is model-free, this is the only information it can use to make decisions. As a result, the RL policy is effectively ``blind'' to any agent that is not immediately in front of it, until that agent actually appears. Because of this, the RL policy cannot anticipate, for example, that an agent would cut in from an adjacent lane; so in order to avoid collisions, 
it converges on more conservative behavior overall.

In contrast, MCTS is model-based and, in our case, it can use the predictions to evaluate the consequences of its actions. As a result, despite using the same low-dimensional state space, MCTS can anticipate cut-ins and other kinds of interactions. For example, if a cut-in is unlikely, it will evaluate actions that make progress as more rewarding; conversely, if a cut-in is likely, it will prefer conservative actions that avoid collisions. This leads to improved progress and safety compared to the standalone RL and IDM planners.

Given the latency costs of the learned policy $\pi_\theta$ and its overly conservative bias (as well as that of $\hat{V}_\theta$), we use the following parameters for the MCTS generator in subsequent experiments: $n = 400, k = 100, P \equiv \mathcal{U}_\mathcal{A}, \pi_\text{rollout} \equiv \pi_\text{IDM}, \pi_\text{padding} \equiv \pi_\text{IDM}$.

\subsection{TreeIRL evaluation}

\begin{table*}[htbp]
\centering
\caption{nuPlan evaluation (7,051 scenarios)}
\label{tab:nuplan}
\begin{tabular}{llccccccc}
\toprule
Category & Metric & IDM & MCTS & PBP & DriveIRL & Gigaflow & DriveIRL-Safe & TreeIRL \\
\midrule
\multirow{5}{*}{Safety} & Collisions $\downarrow$ & 0.02 & \textbf{0.01} & 0.05 & \textbf{0.01} & \textbf{0.01} & \textbf{0.01} & \textbf{0.01} \\
 & Drivable area violations $\downarrow$ & \textbf{0.01} & \textbf{0.01} & 0.03 & \textbf{0.01} & 0.11 & \textbf{0.01} & \textbf{0.01} \\
 & Traffic light violations $\downarrow$ & \textbf{0.02} & \textbf{0.02} & 0.05 & \textbf{0.02} & 0.04 & \textbf{0.02} & \textbf{0.02} \\
 & Speed limit violations $\downarrow$ & \textbf{0.09} & 0.28 & 0.18 & 0.12 & 0.29 & 0.12 & 0.20 \\
 & Time gap (s) $\uparrow$ & 3.73 & 3.68 & 3.63 & 3.85 & 3.42 & \textbf{3.97} & 3.90 \\
\midrule
\multirow{1}{*}{Progress} & Progress along expert route $\uparrow$ & 0.90 & \textbf{0.96} & 0.92 & 0.90 & 0.89 & 0.89 & 0.92 \\
\midrule
\multirow{5}{*}{Comfort} & Comfort $\uparrow$ & 0.92 & 0.98 & 0.97 & \textbf{0.99} & 0.38 & 0.93 & 0.98 \\
 & Min longitudinal jerk (m/s$^3$) $\uparrow$ & -1.08 & -0.87 & -0.66 & \textbf{-0.44} & -2.59 & -0.61 & -0.77 \\
 & Max longitudinal jerk (m/s$^3$) $\downarrow$ & 1.06 & 1.16 & 0.88 & \textbf{0.55} & 2.58 & 0.71 & 0.84 \\
 & Min longitudinal accel (m/s$^2$) $\uparrow$ & -0.99 & -0.67 & -0.52 & \textbf{-0.43} & -1.65 & -0.47 & -0.59 \\
 & Max longitudinal accel (m/s$^2$) $\downarrow$ & 0.41 & 0.75 & 0.67 & \textbf{0.38} & 1.74 & 0.55 & 0.57 \\
\midrule
\multirow{4}{*}{Human-likeness} & L2 error (m) $\downarrow$ & 4.53 & 3.82 & 3.86 & \textbf{3.53} & 8.44 & 3.56 & 3.75 \\
 & Deceleration delay error (s) $\mid\downarrow\mid$ & 0.32 & 0.31 & 0.42 & \textbf{0.14} & 0.78 & 0.15 & 0.43 \\
 & Acceleration delay error (s) $\mid\downarrow\mid$ & \textbf{0.06} & -0.23 & -0.56 & 0.09 & -0.41 & 0.23 & -0.11 \\
 & Max speed error $\downarrow$ & 0.91 & \textbf{0.27} & 0.45 & 1.25 & 0.43 & 1.17 & 0.45 \\
\bottomrule
\end{tabular}
\end{table*}

\begin{table}[htbp]
\centering
\caption{nuPlan evaluation on challenging cut-ins (134 scenarios)}
\label{tab:nuplan_hard_cutins}
\begingroup
\setlength{\tabcolsep}{3pt}
\renewcommand{\arraystretch}{0.85}
\footnotesize
\begin{tabular}{@{}lcccc@{}}
\toprule
Metric & MCTS & DriveIRL & DriveIRL-Safe & TreeIRL \\
\midrule
Collisions $\downarrow$ & 0.31 & 0.34 & \textbf{0.16} & 0.19 \\
Drivable area violations $\downarrow$ & \textbf{0.01} & \textbf{0.01} & \textbf{0.01} & \textbf{0.01} \\
Traffic light violations $\downarrow$ & \textbf{0.00} & 0.01 & 0.01 & \textbf{0.00} \\
Speed limit violations $\downarrow$ & 0.17 & 0.16 & 0.14 & \textbf{0.13} \\
Time gap (s) $\uparrow$ & 0.72 & 0.67 & \textbf{1.02} & 1.01 \\
Progress along expert route $\uparrow$ & \textbf{0.99} & 0.98 & 0.98 & 0.98 \\
Comfort $\uparrow$ & 0.84 & 0.86 & 0.75 & \textbf{0.87} \\
L2 error (m) $\downarrow$ & 6.60 & 5.50 & \textbf{3.30} & 5.28 \\
\bottomrule
\end{tabular}
\endgroup
\end{table}

\begin{figure*}
    \centering
    \includegraphics[width=1\textwidth,trim={0 50 10 0},clip]{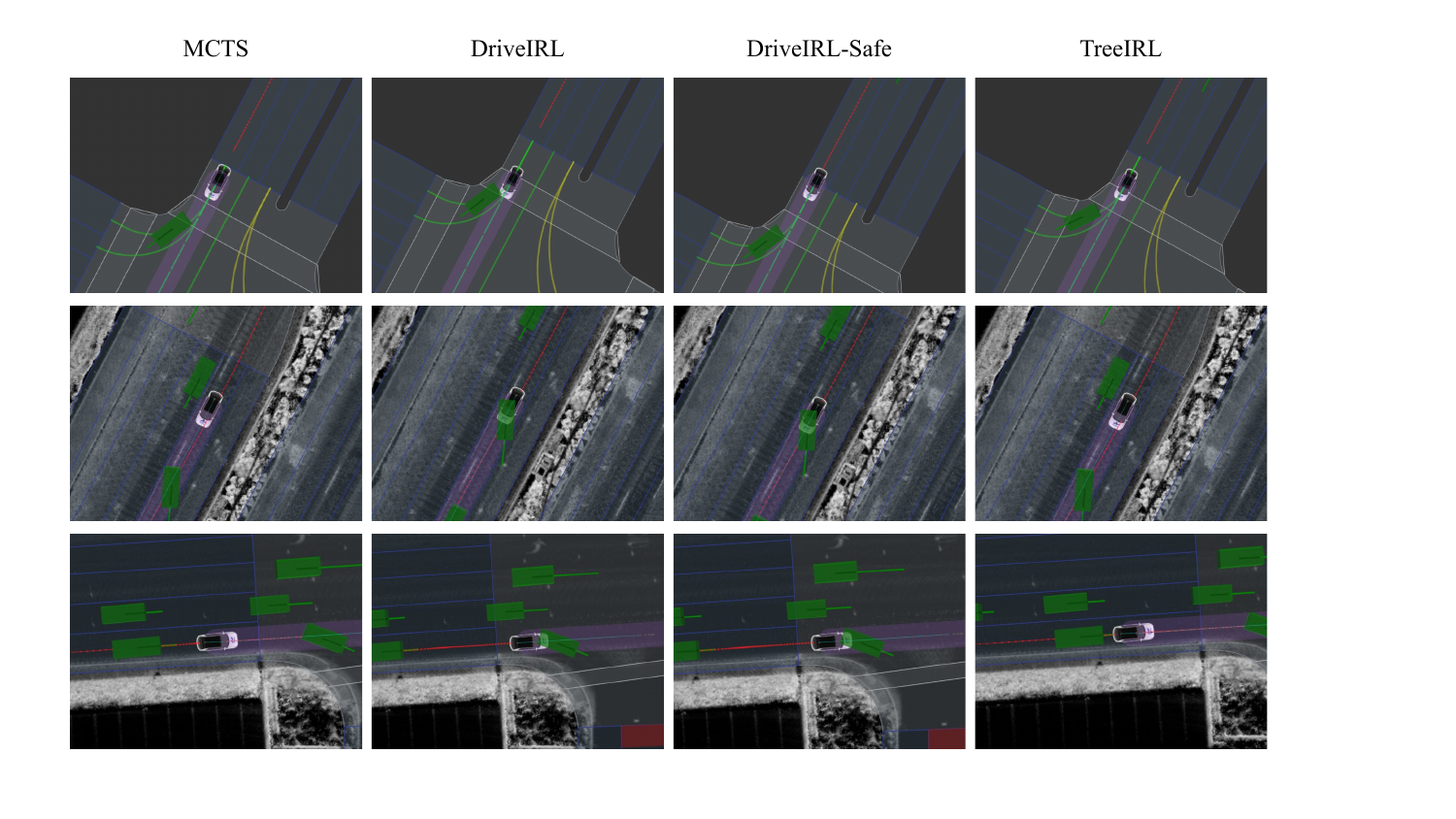}
    \caption{NuPlan examples of DriveIRL collisions avoided by TreeIRL. Each row corresponds to a different scenario. Columns correspond to snapshots of the same moment in time for different planners.}
    \label{fig:nuplan_examples}
\vspace{-3mm}
\end{figure*}


\textbf{Baselines}. We evaluate TreeIRL against the following following classical and state-of-the-art baselines (Fig.~\ref{fig:intro}):

\begin{itemize}
    \item The intelligent driver model (\textbf{IDM})~\cite{treiber2000congested}: a classical planner that computes the acceleration for collision-free ACC in closed form,
    \item \textbf{MCTS}, corresponding to TreeIRL with $k = 1$ trajectory, i.e. treating the IRL scorer as pass-through (Fig.~\ref{fig:arch}, bottom left),
    \item Path-based prediction (\textbf{PBP})~\cite{afshar2023pbp}: an open-loop motion forecasting model trained using imitation learning (Fig.~\ref{fig:arch}, middle left),
    \item \textbf{DriveIRL}~\cite{phan2023driveirl}: precursor to TreeIRL using the same planner architecture, except with a heuristic trajectory generator that generates jerk-optimal trajectories reaching a set of predefined target points (Fig.~\ref{fig:arch}, bottom middle),
    \item \textbf{Gigaflow}~\cite{cusumano2025robust}: a RL-based motion planner trained in closed loop using self-play, with the same policy controlling the ego and all other agents (we use an in-house implementation based on~\cite{cusumano2025robust}, as there is no code/weights available),
    \item \textbf{DriveIRL-Safe}~\cite{phan2023driveirl}: identical to DriveIRL, with the addition of a safety filter as described in~\cite{phan2023driveirl,tomov2023safety} that excludes any apparently unsafe trajectories (Fig.~\ref{fig:arch}, bottom middle).
\end{itemize}

\textbf{Results}. Most planners exhibit comparable and acceptable safety and progress (Tab.~\ref{tab:nuplan}, Fig.~\ref{fig:nuplan_examples}), with the exception of IDM, PBP (collisions), and Gigaflow (drivable area violations). Comfort is comparable across planners, except for Gigaflow. IDM comfort is also on the lower side, especially longitudinal jerk and acceleration. Similarity to the human expert driver is comparable across planners, except for Gigaflow and IDM.

Based on these results, we exclude IDM, PBP, and Gigaflow from subsequent experiments and analyses, focusing only on MCTS, DriveIRL, DriveIRL-Safe, and TreeIRL.

Since the collision metric appears saturated across all scenarios, we further zoom in on a subset of challenging cut-in cases where an agent appears abruptly at a short distance in front of the ego (Tab.~\ref{tab:nuplan_hard_cutins}). On this set, DriveIRL-Safe and TreeIRL have around half as many collisions as MCTS and DriveIRL, while maintaining the same level of progress. While safety and similarity to the human expert driver is better for DriveIRL-Safe compared to TreeIRL, comfort is substantially worse.

\section{Real-world driving}

The ultimate test for a planner is how it performs in the real world. To that end, we deploy the four most promising planners on Hyundar IONIQ 5 self-driving cars and evaluate them on public roads in Las Vegas. 


\textbf{High-fidelity simulation}.
As in previous work~\cite{heim2025lab2car}, prior to on-road deployment, performance of the planning system is thoroughly evaluated using the Object Sim simulator from Applied Intuition \cite{simian}, which performs realistic high-fidelity physics simulation of vehicle dynamics. Unlike nuPlan, which is Python-based, only simulates the planner, and uses a simplified simulator, Object Sim simulates the entire AV stack in a more realistic setting closely resembling real-world deployment. This allows us to get a better sense of on-road performance and to catch integration issues that may not appear in nuPlan. As in nuPlan, other agents are replayed from the log. All simulations are in closed loop, last 30 s (including 4 s of warmup), and run at 10 Hz. 

Overall, nuPlan can be viewed as a cheap and fast way to get an initial sense of planner performance, making it useful for parameter tuning and for ruling out obviously inferior planners. In contrast, Object Sim is a robust yet computationally intensive way to get a more realistic sense of planner performance in the context of a full AV stack.

\textbf{Simulation scenarios}. We use a set of 600+ handpicked scenarios based on on-road events in Las Vegas metropolitan area that were particularly challenging for the AV, including close ACC (5\%), encroachments (2\%), cut-ins (2\%), traffic lights (7\%), low speed (6\%), high speed (11\%), far gap behind lead (15\%), harsh braking (17\%), lane biasing (6\%), jerky driving (24\%). 

\textbf{On-road scenarios}. Routes cover similar geolocations to the simulations, including the Strip, downtown, and west of Strip areas. 
Driving in Las Vegas involves navigating dense urban traffic, handling traffic lights and intersections, smoothly following and breaking for other drivers, responding safely and comfortably to cut-ins and lead vehicle breaking. All on-road tests are performed with an experienced safety driver ready to take over immediately in case of unsafe driving, as well as an experienced test engineer continuously monitoring AV performance.

\textbf{Simulation metrics}. We use a similar set of metrics to nuPlan, with some differences largely due to the differences in the simulators and the data.

\begin{itemize}
    \item \textbf{Front collisions}: instances when the ego collides at the front while moving (speed $>0.01$ m/s). 
    \item \textbf{Traffic light violations}: instances when the ego fully crosses the stop line of a red traffic light.
    \item \textbf{Stop line violations}: instances when the ego partially crosses the stop line of a red traffic light.  
    \item \textbf{Speed limit violation time}: total time spent driving above the speed limit.  
    \item \textbf{ACC distance violations}: instances when the following distance to the lead vehicle drops below 1.5 m.  
    \item \textbf{Min time gap}: minimum time gap to the lead vehicle. Small numbers indicate close calls.
    \item \textbf{Average speed}: average ego speed.  
    \item \textbf{Rear collisions}: instances when the ego is hit from behind. Due to the non-reactive simulation, these are interpreted as a measure of progress: rear collisions indicate slower driving / not keeping with the flow of traffic.  
    \item \textbf{Brake taps}: instances when the negative longitudinal jerk (brake) exceeds $4.25~\text{m/s}^3$.  
    \item \textbf{Longitudinal jerk violations}: instances when the absolute longitudinal jerk exceeds $5~\text{m/s}^3$.  
    \item \textbf{Lateral jerk violations}: instances when the absolute lateral jerk exceeds $7~\text{m/s}^3$.  
\end{itemize}

As in nuPlan, simulation metrics are reported as averages across simulations.

\textbf{On-road metrics}.
Metrics that are useful in simulation may not be practical when assessing real-world driving. For example, collisions will always be 0, as the safety driver will take over before any real collision actually occurs. We therefore measure on-road performance with a combination of discretionary and objective metrics. Discretionary metrics correspond to events manually tagged at the discretion of the test team onboard the AV during road tests. These include takeovers by the safety driver when AV behavior is deemed unsafe, as well as manually flagged events corresponding to comfort or progress issues. Objective metrics are computed after-the-fact based on data logged by the AV.


We use the following discretionary metrics:

\begin{itemize}
    \item \textbf{Safety takeovers}: any takeover by the safety driver due to unsafe AV behavior.
    \item \textbf{ACC failures}: safety takeover preventing collisions with a lead vehicle.
    \item \textbf{Traffic light failures}: safety takeovers preventing traffic light violations.
    \item \textbf{Cut-in failures}: safety takeovers preventing collisions due to cut-ins.
    \item \textbf{Slow driving}: instances when the AV drives too slowly.
    \item \textbf{Discomfort}: any instance of uncomfortable driving (jerky or discomfort braking).
    \item \textbf{Jerky driving}: instance of jerky driving, e.g. brake taps or jerky throttle.
    \item \textbf{Harsh/mild discomfort braking}: instances of severe/light uncomfortable braking.
\end{itemize}

We use the following objective metrics:

\begin{itemize}
    \item \textbf{Harsh/mild discomfort braking}: output of classifier trained to predict human-annotated harsh/mild discomfort braking based on vehicle kinematics.
    \item \textbf{Brake taps}: identical to brake taps in simulation.
\end{itemize}

All on-road metrics are reported as autonomous driving miles (auto-miles/auto-mi) per event.

\subsection{Deployment strategy}

\begin{table}[t]
\centering
\caption{Deployment strategy latency measurements}
\begin{tabular}{lcc}
\toprule
Statistic & DriveIRL + Lab2Car (ms) & DriveIRL + Smoother (ms) \\
\midrule
Max      & 246.70 & 26.16 \\
P99.99   & 246.70 & 26.16 \\
P99      & 208.81 & 20.95 \\
P50      & 60.23  & 14.29 \\
Average  & 73.09  & 14.63 \\
\bottomrule
\end{tabular}
\label{tab:latency_lab2car}
\end{table}

\begin{table*}[htbp]
\centering
\caption{Comparing deployment strategies}
\label{tab:lab2car_vs_smoother}
\begin{tabular}{llcc@{\hskip 6pt}cc}
\toprule
Category & Metric & \multicolumn{2}{c}{DriveIRL} & \multicolumn{2}{c}{MCTS} \\
\addlinespace[0.5ex] 
& & + Lab2Car & + Smoother & + Lab2Car & + Smoother \\
\cmidrule(lr){3-4}\cmidrule(lr){5-6}
\multicolumn{6}{c}{\textbf{High-fidelity 30-s simulations}} \\

\midrule

& Total simulations & 605 & 605 & 605 & 605 \\
\cmidrule(lr){1-2}\cmidrule(lr){3-4}\cmidrule(lr){5-6}
\multirow{4}{*}{Safety}
& Front collisions $\downarrow$ & \textbf{0.05} & \textbf{0.05} & \textbf{0.05} & \textbf{0.05} \\
& Speed limit violation time (s) $\downarrow$ & \textbf{4.14} & 5.76 & 3.94 & \textbf{3.68} \\
& ACC distance violations ($<1.5$ m) $\downarrow$ & \textbf{0.20}	& 0.25	& \textbf{0.19} & 	\textbf{0.19} \\
& Min time gap (s) $\uparrow$ & \textbf{1.26} & 1.01 & \textbf{1.48} & 1.43 \\
\cmidrule(lr){1-2}\cmidrule(lr){3-4}\cmidrule(lr){5-6}
\multirow{1}{*}{Progress}
& Average speed (m/s) $\uparrow$ & \textbf{8.32} & 8.23 & \textbf{9.82} & 9.79 \\
\cmidrule(lr){1-2}\cmidrule(lr){3-4}\cmidrule(lr){5-6}
\multirow{3}{*}{Comfort}
& Brake taps $\downarrow$ & \textbf{0.18} & 0.29 & 0.29 & \textbf{0.27} \\
& Longitudinal jerk violations $\downarrow$ & \textbf{0.12} & 0.17 & 0.29 & \textbf{0.23} \\
& Longitudinal accel violations $\downarrow$ & \textbf{0.23} & 0.29 & 0.54 & \textbf{0.50} \\

\midrule

\multicolumn{6}{c}{\textbf{Public road evaluation}} \\
\cmidrule(lr){1-2}\cmidrule(lr){3-4}\cmidrule(lr){5-6}
& Total auto-miles & 40.8 & 34.8 & 62.1 & 64.1 \\
\cmidrule(lr){1-2}\cmidrule(lr){3-4}\cmidrule(lr){5-6}
\multirow{3}{*}{Comfort}
& Harsh discomfort braking (mi/event) $\uparrow$ & \textbf{40.8} & 8.70 & 5.17 & \textbf{9.16} \\
& Mild discomfort braking (mi/event) $\uparrow$ & \textbf{40.8} & 11.60 & 5.17 & \textbf{8.01} \\
& Brake taps (mi/event) $\uparrow$ & \textbf{1.57} & 0.87 & 0.60 & \textbf{1.49} \\
\bottomrule
\end{tabular}
\begin{minipage}{0.73\textwidth}
\vspace{1mm}
\scriptsize
Deployment strategies compared separately for DriveIRL and MCTS.
\end{minipage}
\end{table*}

Like most ML-based planners, DriveIRL and DriveIRL-Safe do not ensure kinematic feasibility. MCTS and TreeIRL ensure kinematic feasibility only longitudinally. Neither planner ensures dynamic feasibility. This necessitates performing some kind of post-processing on the output trajectory $\hat{\tau}$ before passing it to the downstream drive-by-wire system for execution. 

We consider two deployment strategies:

\begin{itemize}
    \item Lab2Car~\cite{heim2025lab2car}: a two-step process that converts $\hat{\tau}$ into a set of spatiotemporal constraints -- a ``maneuver'' -- that incorporate information about vehicle dynamics, road geometry, and (optionally) other agents (although we do not take advantage of that feature). The resulting optimization problem 
    is solved using an industry-grade MPC solver.
    \item Smoother: a lightweight MPC that applies minor kinematic corrections to $\hat{\tau}$  and performs bookkeeping to ensure continuity between planning cycles. The optimization is performed using a kinematic bicycle model in Cartesian 2D coordinates.
\end{itemize}

We first deploy DriveIRL with each strategy and measure the system latency using open-loop resimulations on identical logs using identical hardware as on the car (Tab.~\ref{tab:latency_lab2car}). Using the Smoother is about an order of magnitude faster than using Lab2Car. However, subsequent closed-loop simulations and real-world driving (Tab.~\ref{tab:lab2car_vs_smoother}) show that DriveIRL + Lab2Car performs better across all metrics compared to DriveIRL + Smoother. Interestingly, we observe the opposite pattern for MCTS.

\textbf{Interim discussion}. We suspect MCTS fails to benefit from Lab2Car due to its bookkeeping mechanism and jerk controls, which make it sensitive to latency, particularly in an asynchronous environment. For example, consider the case when MCTS commands braking with jerk -2 m/s$^3$ from cruising speed. Initially, the deceleration will be small; this will be sent to Lab2Car, which will command a gentle brake. However, while Lab2Car is still running, MCTS (which is must faster and hence runs at higher frequency) will run for several iterations, each commanding jerk -2 m/s$^3$ and assuming that the previous jerk command was executed (due to bookkeeping). MCTS will thus quickly reach the max decel of -7 m/s$^2$. At the next cycle when Lab2Car finally catches up, the MCTS trajectory will now appear to suddenly command a harsh brake, leading to discomfort. 

In contrast, DriveIRL plans from the measured pose, making it more robust to downstream delays. However, this means it may also be less reactive. 

For all subsequent experiments, we deploy DriveIRL/DriveIRL-Safe with Lab2Car and MCTS/TreeIRL with the Smoother.

\subsection{TreeIRL evaluation}

\begin{table*}[htbp]
\centering
\caption{High-fidelity simulation and on-road evaluation}
\label{tab:simian_and_road}
\begin{tabular}{llcccc}
\toprule
Category & Metric & MCTS & DriveIRL & DriveIRL-Safe & TreeIRL \\
\midrule
\multicolumn{6}{c}{\textbf{High-fidelity 30-s simulations}} \\
\midrule
& Total simulations & 717 & 717 & 717 & 717 \\
\midrule
\multirow{6}{*}{Safety}
& Front collisions $\downarrow$ & 0.04 & 0.05 & \textbf{0.03} & \textbf{0.03} \\
& Traffic light violations $\downarrow$ & \textbf{0.01} & 0.04 & 0.04 & \textbf{0.01} \\
& Stop line violations $\downarrow$ & \textbf{0.01} & 0.06 & 0.04 & \textbf{0.01} \\
& Speed limit violation time (s) $\downarrow$ & 3.81 & 4.39 & \textbf{2.07} & 5.01 \\
& ACC distance violations ($<1.5$ m) $\downarrow$ & \textbf{0.17} & 0.20 & 0.24 & 0.19 \\
& Min time gap (s) $\uparrow$ & \textbf{1.56} & 1.29 & 1.31 & 1.45 \\
\midrule
\multirow{2}{*}{Progress}
& Average speed (m/s) $\uparrow$ & \textbf{8.88} & 8.41 & 8.31 & \textbf{8.88} \\
& Rear collisions $\downarrow$ & 0.17 & \textbf{0.16} & 0.28 & 0.19 \\
\midrule
\multirow{3}{*}{Comfort}
& Brake taps $\downarrow$ & 0.27 & \textbf{0.18} & 0.32 & 0.22 \\
& Longitudinal jerk violations $\downarrow$ & 0.23 & \textbf{0.12} & 0.23 & 0.17 \\
& Longitudinal accel violations $\downarrow$ & 0.49 & \textbf{0.23} & 0.31 & 0.56 \\
\midrule
\multicolumn{6}{c}{\textbf{Public road evaluation}} \\
\midrule
& Total auto-miles & 115.8 & 64.3 & 87.9 & 268.4 \\
\midrule
\multicolumn{6}{c}{Discretionary metrics} \\
\midrule
\multirow{4}{*}{Safety}
& Safety takeovers (mi/event) $\uparrow$ & 7.68 & 1.43 & 6.76 & \textbf{17.89} \\
& ACC failures (mi/event) $\uparrow$ & 57.9 & 2.57 & 87.9 & \textbf{$>$268.4} \\
& Traffic light failures (mi/event) $\uparrow$ & 19.3 & 8.04 & 12.56 & \textbf{67.10} \\
& Cut-in failures (mi/event) $\uparrow$ & \textbf{$>$115.8} & 5.36 & 43.95 & \textbf{$>$268.4} \\
\midrule
\multirow{1}{*}{Progress}
& Slow driving (mi/event) $\uparrow$ & \textbf{$>$115.8} & 5.85 & 2.84 & \textbf{134.2} \\
\midrule
\multirow{4}{*}{Comfort}
& Discomfort (mi/event) $\uparrow$ & 1.40 & 1.74 & 1.05 & \textbf{2.42} \\
& Jerky driving (mi/event) $\uparrow$ & 2.83 & 6.43 & 2.20 & \textbf{13.42} \\
& Harsh discomfort braking (mi/event) $\uparrow$ & 6.09 & 4.95 & 3.66 & \textbf{8.95} \\
& Mild discomfort braking (mi/event) $\uparrow$ & 2.83 & 3.22 & 3.26 & \textbf{12.78} \\
\midrule
\multicolumn{6}{c}{Objective metrics} \\
\midrule
\multirow{3}{*}{Comfort}
& Harsh discomfort braking (mi/event) $\uparrow$ & 8.91 & 4.29 & 4.88 & \textbf{9.59} \\
& Mild discomfort braking (mi/event) $\uparrow$ & 7.24 & \textbf{64.3} & 17.58 & 15.79 \\
& Brake taps (mi/event) $\uparrow$ & 1.08 & 0.43 & 1.03 & \textbf{1.11} \\
\bottomrule
\end{tabular}
\end{table*}

\begin{figure*}
    \centering
    \includegraphics[width=1\textwidth,trim={0 10 0 0},clip]{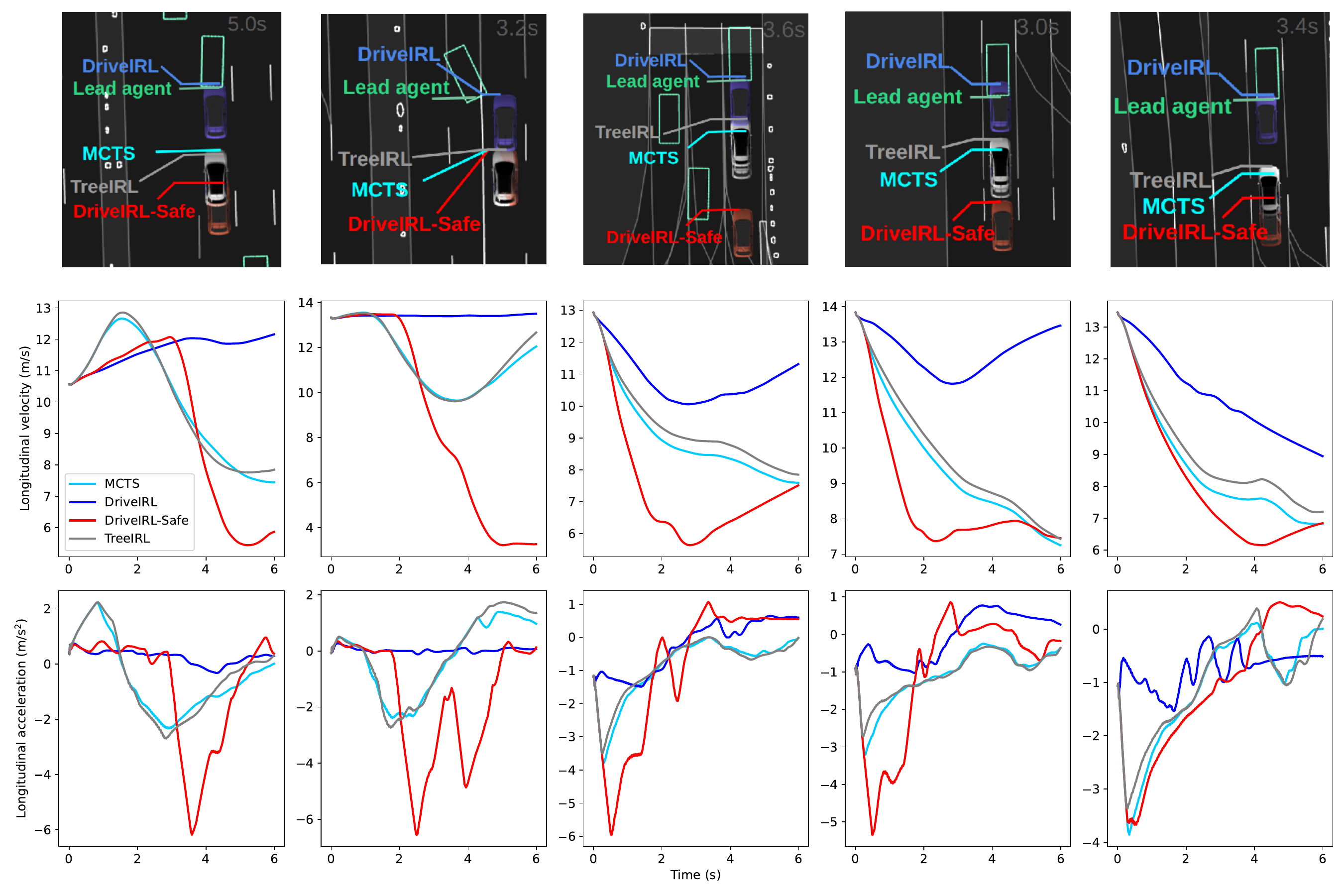}
    \caption{High-fidelity simulation examples. Each column corresponds to a different scenario. Top row, snapshot of the same moment in time in the simulation for each planner.}
    \label{fig:simian_examples}
\vspace{-3mm}
\end{figure*}

\begin{figure*}
    \centering
    \includegraphics[width=1\textwidth,trim={0 50 10 0},clip]{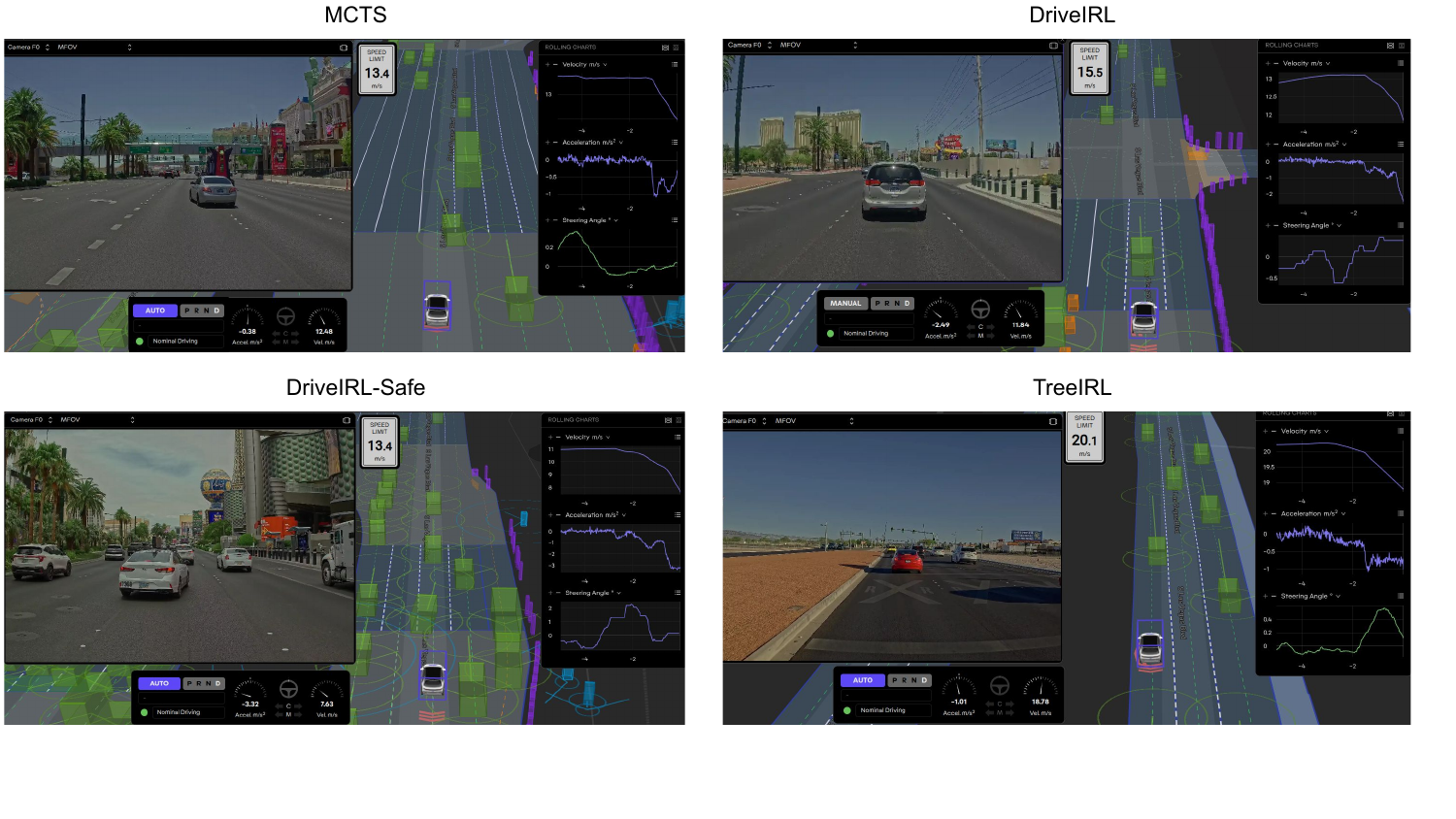}
    \caption{Cut-in examples from real-world driving. Right inset, recent history of AV kinematics. For DriveIRL, $t = 0$ corresponds to a safety-related takeover (cut-in failure).}
    \label{fig:real}
\vspace{-3mm}
\end{figure*}

We integrate MCTS, DriveIRL, DriveIRL-Safe, and TreeIRL with the Motional AV-stack (Fig.~\ref{fig:arch}) and compare them in simulation, using our full-stack simulator, and in the real world, using Hyundai IONIQ 5 self-driving cars (Tab.~\ref{tab:simian_and_road}, Fig.~\ref{fig:simian_examples} \& \ref{fig:real}).

High-fidelity simulations largely recapitulate the nuPlan results, with DriveIRL showing better comfort and TreeIRL showing better safety. TreeIRL generally shows improved comfort over MCTS while having relatively comparable safety and progress, highlighting the benefit of the IRL scorer. DriveIRL-Safe has comparable safety to TreeIRL (except for traffic light and stop line violations), but worse comfort overall. Analyses of individual simulations (Fig.~\ref{fig:simian_examples}) show that DriveIRL fails to stop on time to prevent collisions, while DriveIRL-Safe stops too soon and too abruptly. Together, these results validate the conclusions from the nuPlan simulations and the choice of deployment strategy. 

The on-road results, however, overwhelmingly favor TreeIRL across all metrics. Safety is $\sim$2 orders of magnitude better than DriveIRL and 2-4x better than DriveIRL-Safe and MCTS, with zero takeovers by the safety driver due to ACC or cut-in failures across all 268 auto-miles. Progress is comparable to MCTS and substantially better than DriveIRL and DriveIRL-Safe. Analysis of individual on-road cut-ins (Fig.~\ref{fig:real}) reveal similar kinematic profiles to the simulations: MCTS deceleration is somewhat jerky, DriveIRL slows down insufficiently to prevent takeovers, DriveIRL-Safe brakes too aggressively, and TreeIRL slows down with the smoothest deceleration profile. Interestingly, in contrast to the simulations, both subjective and objective comfort is better for TreeIRL compared to DriveIRL. 

\textbf{Interim discussion}. The substantially larger advantage of TreeIRL over the other planners on the road compared to simulation is partly due to the subjective nature of the discretionary metrics. Specifically, the safety driver may take over if AV behavior is perceived to be unsafe, even if it would not have resulted in a collision. Indeed, resimulations confirm that this is the case for a substantial number of takeovers. However, as a measure of perceived safety, takeovers better reflect the subjective experience of the passenger, which in some sense is the ultimate yardstick for measuring AV performance. The sudden braking during takeovers additionally impacts comfort, which is likely the main factor behind the worse comfort of DriveIRL on the road compared to the simulations.

It is worth noting that the on-road results for DriveIRL and DriveIRL-Safe are in line with previous reports~\cite{phan2023driveirl} of 2 takeovers for 8.8 auto-miles and 0 takeovers for 6.9 auto-miles, respectively, albeit on a much smaller evaluation.

\section{General discussion and conclusions}

We presented TreeIRL, a novel combination of MCTS and IRL for autonomous driving that outperforms the state of the art in simulated and real-world urban driving in terms of safety, progress, comfort, and human-likeness. To the best of our knowledge, this is the first real-world evaluation of a motion planner based on MCTS. We suspect this is due to latency: since the size of the search space (i.e., number of possible trajectories) is exponential in the size of the state space, most applications of MCTS to motion planning require more iterations to converge than is practically feasible on a real AV. One solution is to use a learned policy~\cite{hoel2019combining}; however, this requires performing inference-in-the-MCTS-loop -- potentially multiple times during the rollouts -- which further strains latency. While this can be overcome with more compute and parallelization~\cite{silver2016mastering}, this is not always feasible on board a real AV where multiple systems compete for scarce resources in real time.

Instead, by combining MCTS with IRL, we can relax the demands on the MCTS generator, as it no longer needs to find the single best trajectory but merely to home in on a promising subset of the trajectory space and lean on the IRL scorer to choose the best trajectory. One way to view this is that MCTS handles safety, by choosing a safe behavior mode, while IRL handles comfort, by choosing the most human-like variation around that mode. Indeed, our simulation (Tab.~\ref{tab:nuplan}) and on-road (Tab.~\ref{tab:simian_and_road}) results indicate that the biggest gain of TreeIRL over vanilla MCTS is in comfort, although safety is also improved.


On the road, TreeIRL showed close to 2 orders of magnitude better safety than DriveIRL (Tab.~\ref{tab:simian_and_road}), highlighting the benefits of combining classical and learning-based approaches. Accordingly, the advantage of TreeIRL over DriveIRL-Safe -- another hybrid system -- was smaller. However, DriveIRL-Safe was inferior to both TreeIRL and DriveIRL in terms of comfort and progress. This highlights the limitations of na\"ive enumerative trajectory generation, which can fail to provide sufficient coverage of the trajectory space: once the safety filter rules out apparently ``bad'' trajectories, there may be insufficient diversity in the remaining trajectories to ensure comfort and progress. 

The emergence of public simulators and benchmarks such as nuPlan~\cite{karnchanachari2024towards}, CARLA~\cite{dosovitskiy2017carla}, and Waymax~\cite{ettinger2021large,gulino2023waymax} has been instrumental in advancing motion planning research by facilitating the comparison of different planners on an equal footing. However, compressing planner performance into a single scalar score can obfuscate critical planner limitations (Tab.~\ref{tab:nuplan}). Until a single universal gold standard metric is designed and agreed upon, we believe planners should be compared holistically across a diverse set of metrics capturing safety, progress, and comfort (Fig.~\ref{fig:intro}, bottom right). 

Our work also highlights the limitations of relying solely on simulation to assess planner performance. The discrepancy between simulated and real-world performance -- the sim-to-real gap -- is well-documented~\cite{tobin2017domain,kiran2021deep}, yet it is nevertheless striking that the relatively small advantage of TreeIRL in simulation (Tab.~\ref{tab:nuplan}) translated to 1-2 orders of magnitude improvement on the road (Tab.~\ref{tab:simian_and_road}). Until simulation fidelity improves to reliably match on-road performance -- a problem nearly as hard as the planning problem itself -- we believe early on-road tests~\cite{heim2025lab2car} should be a critical part of comprehensive planner evaluation. Thus planner comparison should proceed both holistically and incrementally, with a broad set of experimental planners evaluated in simulation on a broad set of metrics and then narrowed down to a smaller set of promising planners that are deployed and evaluated in the real world.

An alternative way to combine MCTS with IRL would be to replace the handcrafted reward function $R$ (Eq.~\ref{eq:jerk_cost}-\ref{eq:station_buffer_cost}) with a reward function learned from data. This learned reward function can be directly plugged into MCTS and/or used to train the RL network $f_\theta$. This could obviate the need for a learned trajectory scorer, as the reward function itself would already capture human-likeness. Yet another way to rid of the IRL scorer would be to train $f_\theta$ with a hybrid of RL and IL~\cite{lu2023imitation}. 

Our work can be further extended in multiple ways. Prediction uncertainty could be accounted for by considering multiple prediction modes and averaging rewards over them. Alternatively, predictions could be replaced by a reactive world model, such as TrafficSim~\cite{suo2021trafficsim} or Gigaflow~\cite{cusumano2025robust}. Gigaflow could also facilitate the expansion of the state/action space to include lateral control by replacing the RL network $f_\theta$, as planning in 2D instead of 1D would likely require prohibitively many iterations if MCTS is guided by a simple policy such as the IDM. This would enable new capabilities such as lane biasing and lane changes, 
which we leave as the subject of future work. Overall, TreeIRL can be seen as a framework for tackling the planning bottleneck in autonomous driving by combining the strengths of tree search, RL, IL, and IRL in a single planner architecture that facilitates robust comparisons in simulation and in the real world.



\section*{Acknowledgments}

We thank Raveen Ilaalagan, Ernest Elizalde, and Xavier Escobar for road tests; Michael Zahniser, Curtis Chan, Samee Mahbub, and Joseph Baker for deployment/integration; and the countless other Motional scientists and engineers who contributed to this project directly or indirectly.

\bibliographystyle{IEEEtran}
\bibliography{IEEEabrv, references}


\end{document}